\documentclass{article}

\usepackage[preprint]{neurips_2026_local}

\usepackage[utf8]{inputenc}
\usepackage[T1]{fontenc}
\usepackage{hyperref}
\usepackage{url}
\usepackage{booktabs}
\usepackage{array}
\usepackage{amsfonts}
\usepackage{nicefrac}
\usepackage{microtype}
\usepackage{xcolor}
\usepackage{graphicx}
\usepackage{wrapfig}
\usepackage{subcaption}
\usepackage{multirow}
\usepackage{adjustbox}
\usepackage{amsmath}
\usepackage{amssymb}
\usepackage{mathtools}
\usepackage{amsthm}
\usepackage{float}
\usepackage{algorithm}
\usepackage{algpseudocode}
\usepackage[capitalize,noabbrev]{cleveref}
\usepackage[disable,textsize=tiny]{todonotes}

\algrenewcommand\algorithmicrequire{\textbf{Input:}}
\algrenewcommand\algorithmicensure{\textbf{Output:}}
\newcommand{\rev}[1]{#1}

\theoremstyle{plain}
\newtheorem{theorem}{Theorem}[section]

\newtheorem{lemma}[theorem]{Lemma}
\newtheorem{corollary}[theorem]{Corollary}
\theoremstyle{definition}

\newtheorem{assumption}[theorem]{Assumption}
\theoremstyle{remark}

\title{Neural CDEs as Correctors for Learned Time Series Models}

\author{%
  Muhammad Bilal Shahid \\
  Department of Mechanical Engineering \\
  Iowa State University \\
  \texttt{belal@iastate.edu} \\
  \And
  Zhanhong Jiang \\
  Department of Mechanical Engineering \\
  Iowa State University \\
  \texttt{zhjiang@iastate.edu} \\
  \And
  Prajwal Koirala \\
  Sibley School of Mechanical and Aerospace Engineering \\
  Cornell University \\
  \texttt{pk596@cornell.edu} \\
  \And
  Soumik Sarkar \\
  Department of Mechanical Engineering \\
  Iowa State University \\
  \texttt{soumiks@iastate.edu} \\
  \And
  Cody Fleming \\
  Department of Mechanical Engineering \\
  Iowa State University \\
  \texttt{flemingc@iastate.edu} \\
}

\begin{document}

\maketitle

\begin{abstract}

Learned time-series models, whether continuous or discrete, are widely used for forecasting the states of dynamical systems but suffer from error accumulation in multi-step forecasts. To address this issue, we propose a Predictor–Corrector framework in which the Predictor is a learned time-series model that generates multi-step forecasts and the Corrector is a neural controlled differential equation that corrects the forecast errors. The Corrector works with irregular sampled time series and is compatible with both continuous- and discrete-time Predictors. We further introduce two regularization strategies that improve the Corrector's extrapolation performance and accelerate its training. We also provide theoretical guarantees on the stability and convergence of the proposed framework. Experiments on synthetic, physics-based, and real-world datasets show that the proposed framework consistently improves forecasting performance across diverse Predictors, including neural ordinary differential equations, ContiFormer, and DLinear, demonstrating its predictor-agnostic nature.
\end{abstract}

\section{Introduction}

Learned time-series models generate multi-step forecasts either autoregressively, feeding back their own predictions at each step, or non-autoregressively over the full horizon at once \citep{chen2019neuralordinarydifferentialequations,zeng2022transformerseffectivetimeseries}. In both cases, forecast errors accumulate with horizon length, causing predicted trajectories to diverge from ground truth \citep{janner2021offlinereinforcementlearningbig,zhou2022fedformerfrequencyenhanceddecomposed}. The key question is whether these errors contain structure that can be learned to mitigate error accumulation.

\begin{wrapfigure}[18]{r}{0.38\textwidth}
    \vspace{-12pt}
    \centering
    \includegraphics[width=0.95\linewidth]{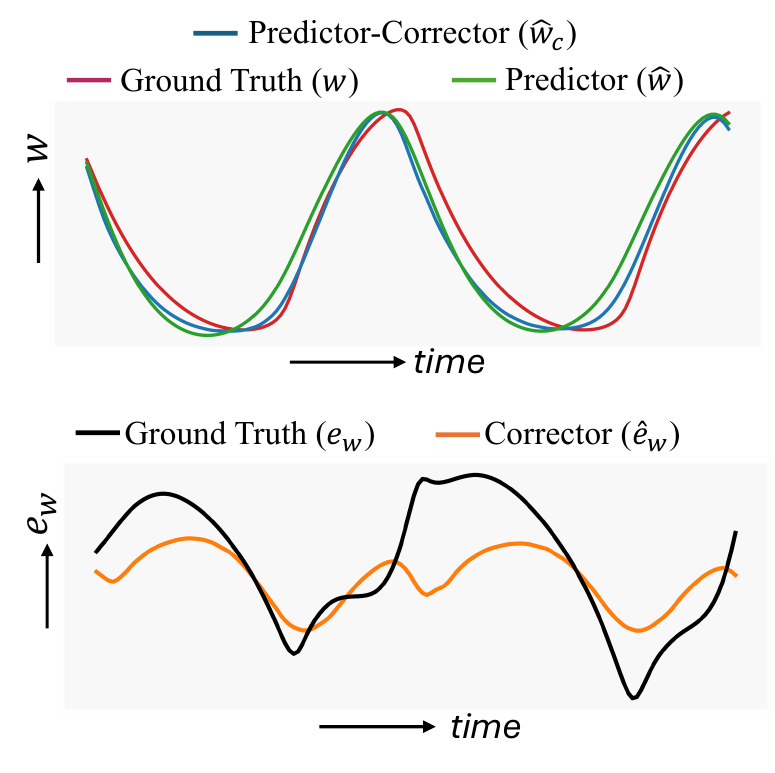}
    \caption{NODE forecasting on \textsc{FHN}. \textbf{(a) Top:} Forecasts with and without Corrector. \textbf{(b) Bottom:} Predicted vs. true error.}
    \label{fig:intro_fig}
\end{wrapfigure}

Fig.~\ref{fig:intro_fig} motivates our approach: a learned time-series model NODE\footnote{Henceforth, \emph{Predictor} refers to any learned time-series model.} \citep{chen2019neuralordinarydifferentialequations} on the FitzHugh-Nagumo (\textsc{FHN}) system \citep{izhikevich2006fitzhugh} produces structured forecast errors (Fig.~\ref{fig:intro_fig}b) that vary with the forecast trajectory. This structure suggests that residuals should remain conditioned on the evolving forecast path, rather than be treated as unstructured noise. We employ a Neural CDE Corrector \citep{kidger2020neuralcontrolleddifferentialequations} that takes Predictor forecasts as a control path and learns path-conditioned residual dynamics ($\hat{e}_w$, $\hat{e}_v$). Adding these to the forecasts yields improved trajectories ($\hat{w}_c$, $\hat{v}_c$) shown in Fig.~\ref{fig:intro_fig}(a). We propose a Predictor-Corrector framework and instantiate it with NODE, ContiFormer \citep{chen2024contiformercontinuoustimetransformerirregular}, and DLinear \citep{zeng2022transformerseffectivetimeseries} as Predictors and Neural CDE as Corrector. Our contributions are:

\begin{itemize}
\item A \textbf{Predictor–Corrector framework} where the Corrector is a Neural CDE modeling residual error dynamics, with \textbf{two control-path regularization strategies} to improve extrapolation and accelerate training, and stability/convergence \textbf{guarantees}.
\item Our framework is \textbf{Predictor-agnostic}: it requires no architectural modifications to the Predictor, is compatible with continuous- and discrete-time Predictors, and
works with regularly or irregularly sampled time series.
\item Consistent improvement in forecast performance across \textbf{synthetic, physics simulation, and LTSF datasets}, demonstrating broad applicability of our proposed framework.
\end{itemize}

\section{Related works}

\paragraph{Predictor-Corrector Framework} We review three uses of Predictor–Corrector ideas in time-series modeling. In forecasting, hybrid models often learn residuals of a base forecaster: ARIMA--MLP \citep{zhang2003time} and ARIMA--LSTM drought models \citep{xu2022application} are early examples, and related forecast-correction ideas appear in recent hybrid forecasting work \citep{chen2022machine,slater2023hybrid,liu2025explainable,zhang2022hybrid}. These methods motivate residual correction but usually do not formalize a predictor-agnostic framework for multi-step forecast trajectories.

Data-assimilation-style approaches also combine prediction and correction \citep{law2015data}, but their correction uses observations or measurements unavailable at forecast time. PhyDNet corrects learned PDE dynamics using observations \citep{guen2020disentangling}; KalmanNet learns Kalman-style gains with GRUs \citep{cho2014properties,revach2022kalmannet}; and NODA extends correction to neural operators with sparse noisy measurements \citep{singh2024learningsemilinearneuraloperators}. Numerical Predictor–Corrector methods instead use explicit and implicit steps to solve differential equations \citep{diethelm2002predictor}, with recent extensions to Transformers \citep{vaswani2017attention,li2024predictorcorrectorenhancedtransformersexponential} and diffusion sampling \citep{zhao2023unipcunifiedpredictorcorrectorframework}. Our setting is distinct: the Corrector operates only on the Predictor's future forecast path, without future observations, and is compatible with continuous- and discrete-time Predictors under regular or irregular sampling.

\paragraph{Continuous-Time Models} Standard RNNs have undefined hidden states between observations, making them poorly suited for irregular sampling. Several approaches model continuous dynamics with exponential decay \citep{che2018recurrent,rajkomar2018scalable} or neural ODEs between observations \citep{rubanova2019latentodesirregularlysampledtime,lechner2020learninglongtermdependenciesirregularlysampled,debrouwer2019gruodebayescontinuousmodelingsporadicallyobserved}. These approaches update continuously \textbf{between} observations after receiving discrete inputs. Neural CDE instead models continuous dynamics \textbf{across} observations through a continuously evolving control path, so the hidden state remains coupled to the whole forecast trajectory. This distinction is central for residual correction, where the error dynamics depends on how the Predictor forecast evolves over the horizon.

\section{Problem description}\label{problem_description}

Consider a $D$-dimensional dynamical system with unknown vector field $\mathbf{f}$, whose trajectory satisfies:
\begin{equation}
    \mathrm{\mathbf{x}}(t) = \mathrm{\mathbf{x}}_{0} + \int_{0}^{t} \mathbf{f}(\mathbf{x}(\tau))d\tau.
    \label{eq:ode_definition}
\end{equation}
We observe $N$ time series of $T$ irregularly sampled observations $\{\mathbf{x}_{i}\}_{i=0}^{T-1}$ at times $t_0 < \cdots < t_{T-1}$, $\mathbf{x}_i \in \mathbb{R}^D$. A Predictor learned from data (e.g., NODE) generates $T$-step forecasts $\{\hat{\mathbf{x}}_{i}\}_{i=0}^{T-1}$, autoregressively or non-autoregressively. The residual error trajectory is $\{\mathbf{e}_{i}\}_{i=0}^{T-1} = \{\mathbf{x}_i - \hat{\mathbf{x}}_i\}_{i=0}^{T-1}$. We hypothesize that a learned vector field $f_{\theta}$ driven by a control path $\mathbf{X}$ built from $\{\hat{\mathbf{x}}_i\}$ can predict these error dynamics $\{\hat{\mathbf{e}}_{i}\}_{i=0}^{T-1}$, yielding corrected forecasts $\{\hat{\mathbf{x}}_i + \hat{\mathbf{e}}_i\}_{i=0}^{T-1}$. This turns residual forecasting into a controlled-dynamics problem, instantiated next with a Neural CDE.

\section{Methodology}

\paragraph{Corrector Design.}\label{sec:corrector_design}

We model the path-conditioned residual process with a Neural CDE \citep{kidger2020neuralcontrolleddifferentialequations}. The Neural CDE is analogous to a continuous-time RNN whose hidden state $\mathbf{z}: [t_0,t_{T-1}] \rightarrow \mathbb{R}^{C}$ evolves continuously under the forecast-derived control path $\mathbf{X}: [t_0,t_{T-1}] \rightarrow \mathbb{R}^{D+1}$. Mathematically,

\begin{equation}
    \mathbf{z}(t) = \mathbf{z}(t_0) + \int_{t_0}^{t} f_{\theta}(\mathbf{z}(s))d\mathbf{X}(s) \,\,\,\, \text{for} \,\,\,\, t \in (t_0,t_{T-1}],
    \label{eq:neural_cde}
\end{equation}

The integral is Riemann-Stieltjes. The control path $\mathbf{X}(s)$ is a cubic Hermite spline with backward differences \citep{morrill2022choice} interpolating $\{(t_i, \hat{\mathbf{x}}_i)\}_{i=0}^{T-1}$ (Appendix~\ref{adx:abl_inter_schemes}), with $\mathbf{X}(t_i) = (\hat{\mathbf{x}}_i, t_i) \in \mathbb{R}^{D+1}$. Thus, the Corrector is conditioned not only on sampled forecasts but also on local increments of the interpolated forecast path. The vector field $f_\theta\colon\mathbb{R}^{C}\to\mathbb{R}^{C\times(D+1)}$ and encoder $\zeta_{\phi}\colon\mathbb{R}^{D+1}\to\mathbb{R}^{C}$ (with $\mathbf{z}(t_0)=\zeta_\phi(\hat{\mathbf{x}}_0, t_0)$) are feedforward networks; ``$f_\theta(\mathbf{z}(s))d\mathbf{X}(s)$'' is a matrix-vector product. A decoder $\xi_\varphi$ maps $\mathbf{z}(t_i)$ to predicted error $\hat{\mathbf{e}}_i$ (Appendix~\ref{adx:decoder_size}).

\begin{figure}[ht]
    \centering
    \includegraphics[width=0.8\columnwidth]{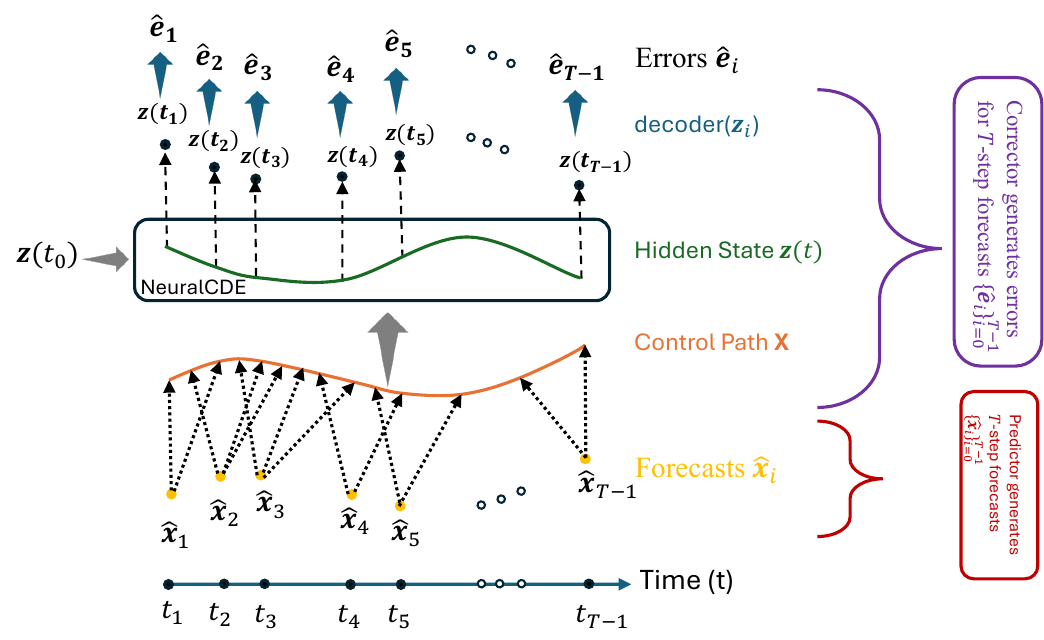}
    \caption{At inference time, Predictor forecasts define a continuous control path that drives the Neural CDE Corrector, whose decoded hidden states provide residual corrections over the forecast horizon.}
    \label{fig:method_fig}
\end{figure}

Algorithm~\ref{alg:pc-train} details training. The Corrector parameters $\theta, \phi, \varphi$ are learned via adjoint-based backpropagation by minimizing:
\begin{equation}
    \mathcal{L} = \frac{1}{NT}\sum_{j=0}^{N-1} \sum_{i=0}^{T-1} ||\mathbf{\hat{e}}_{i,j} - \mathbf{e}_{i,j}||^{2}_{2}
\end{equation}
using $\mathtt{diffrax}$ \citep{kidger2022neuraldifferentialequations} in JAX \citep{jax2018github}; further details are in Appendix~\ref{adx:neural_cde}. At inference, the Predictor forecast path drives the Corrector, which decodes hidden states into residuals and adds them to the original forecasts (Fig.~\ref{fig:method_fig}).

\begin{algorithm}[!htbp]
\caption{Training the Neural CDE as a Corrector}
\label{alg:pc-train}
\begin{algorithmic}[1]
\Require Errors $\{\mathbf{e}_{i}\}$, forecasts $\{\hat{\mathbf{x}}_{i}\}$, times $\{t_i\}$ ($i=0,\dots,T{-}1$) for $N$ trajectories
\Ensure Trained parameters $(\theta, \phi, \varphi)$ of Corrector 
\State Initialize $\theta, \phi, \varphi$
\For{epoch $=1,\dots,E$}
    \State $\mathbf{X}(s) \leftarrow \mathtt{CubicSpline}(\{(t_i,\hat{\mathbf{x}}_{i})\}_{i=0}^{T-1})$ \quad // \emph{Control path}
    \State $\mathbf{z}(t_0) \leftarrow \zeta_{\phi}(\hat{\mathbf{x}}_0,t_{0})$
    \State $\mathbf{z}(t) \leftarrow \mathrm{NeuralCDE}(\mathbf{z}(t_0),\, \mathbf{X}(s),\, f_{\theta})$
    \State $\{\hat{\mathbf{e}}_i\}_{i=1}^{T-1} \leftarrow \{\xi_{\varphi}(\mathbf{z}(t_i))\}_{i=1}^{T-1}$ \quad // \emph{Evaluate $\mathbf{z}(t)$ at $\{t_{i}\}_{i=1}^{T-1}$}
    \State Minimize $\text{MSE}(\mathbf{e}_i,\hat{\mathbf{e}}_i)$ \quad // \emph{mean squared error}
\EndFor
\State \textbf{return} $\theta, \phi, \varphi$
\end{algorithmic}
\end{algorithm}

\subsection{Control paths regularization}\label{sec:reg_strat}

We regularize the control path so the Corrector does not overfit to a single training horizon or sampling density, improving \textit{extrapolation performance} (see Section~\ref{subsec:inter_and_extra_polation}).

\subsubsection{Variable-length control paths \texorpdfstring{$\mathbf{\bar{X}}(s)$}{X-bar(s)}}

During training, the last $k \sim \text{Uniform}\{0,\ldots,\eta\}$ forecasts are dropped, constructing a variable-length path $\bar{\mathbf{X}}(s): [t_0, t_{T-k}] \to \mathbb{R}^{D+1}$ and varying the integration limits in \eqref{eq:neural_cde}:

\begin{equation}\label{eq_5}
    \mathbf{z}(t) = \mathbf{z}(t_0) + \int_{t_0}^{t} f_{\theta}(\mathbf{z}(s))d\mathbf{\bar{X}}(s) \,\,\,\, \text{for} \,\,\,\, t \in (t_0,t_{T-k}],
\end{equation}

\subsubsection{Sparse control paths \texorpdfstring{$\mathbf{\hat{X}}(s)$}{X-hat(s)}}\label{sec:sparse_ctrl}

For synthetic and physics datasets, we simulate irregularly sampled time series at four sparsity levels (20\%, 50\%, 80\%, 100\%) \citep{rubanova2019latentodesirregularlysampledtime}. A second stage retains a fraction $\kappa \in (0,1]$ of the already-sparse points to construct a sparse path $\hat{\mathbf{X}}(s)$, reducing overfitting and improving extrapolation. For LTSF, only $\kappa$-regularization is applied.

\section{Theoretical analysis}
To provide theoretical insight into the proposed Predictor-Corrector framework, we analyze the stability and convergence of the Neural CDE Corrector. All proofs are deferred to Appendix~A. Given the problem formulation in Section~\ref{problem_description}, we start with the assumptions as follows.
\begin{assumption}\label{assumption_1}
    The control path $\mathbf X$ has bounded total variation on any finite interval:
    $
        \|\mathbf{X}\|_{TV,[0,T]}<\infty, \forall T<\infty.
    $
\end{assumption}
\begin{assumption}\label{assumption_2}
    There exists $L_\zeta>0$ such that
    $
        \|\zeta_\phi(\hat{\mathbf{x}}_1)-\zeta_\phi(\hat{\mathbf{x}}_2)\|\leq L_\zeta\|\hat{\mathbf{x}}_1-\hat{\mathbf{x}}_2\|,\forall \hat{\mathbf{x}}_1,\hat{\mathbf{x}}_2
    $
\end{assumption}
\begin{assumption}\label{assumption_3}
    There exists $L_f>0$ such that
    $
        \|f_\theta(\mathbf z_1)-f_\theta(\mathbf z_2)\|\leq L_f\|\mathbf z_1-\mathbf z_2\|,\forall \mathbf z_1,\mathbf z_2.
    $
\end{assumption}
\begin{assumption}\label{assumption_4}
    There exists $L_\xi>0$ such that
    $
        \|\xi_\varphi(\mathbf z_1)-\xi_\varphi(\mathbf z_2)\|\leq L_\xi\|\mathbf z_1-\mathbf z_2\|,\forall \mathbf z_1,\mathbf z_2.
    $
\end{assumption}
The Lipschitz continuity assumptions are standard and mild in neural dynamical systems. In practice, the encoder, Neural CDE vector field, and decoder are implemented as finite-depth neural networks with continuous activations, for which Lipschitz continuity holds under bounded weights and can be enforced via weight decay or spectral normalization. These assumptions ensure well-posedness and robustness of the Neural CDE, and provide the minimal regularity needed to establish stability and convergence guarantees.
To analyze stability and convergence, we consider two Corrector trajectories driven by the same Predictor trajectory. 
Let
$
    \mathbf z_1(t_0)=\zeta_\phi(\hat{\mathbf{x}}_0), \mathbf z_2(t_0)=\zeta_\phi(\hat{\mathbf{x}}'_0)
$
and
$
    \Delta\mathbf z(t):=\mathbf z_1(t)-\mathbf z_2(t),\Delta\hat{\mathbf e}(t):=\hat{\mathbf e}_1(t) - \hat{\mathbf e}_2(t), \Delta\hat{\mathbf{x}}_0=\hat{\mathbf{x}}_0-\hat{\mathbf{x}}'_0.
$
With these in hand, we now first establish the stability as follows.
\begin{theorem}\label{theorem_1}
    Let all assumptions hold. For all $t\geq t_0\geq  0$, we have
    $
        \|\Delta\mathbf z(t)\|\leq L_\zeta\|\Delta\hat{\mathbf{x}}_0\|\text{exp}(L_f\|\mathbf{X}\|_{TV,[0,t]}).
    $
\end{theorem}
Analogously, we have the similar conclusion for the decoder.
\begin{corollary}\label{corollary_1}
    Let all assumptions hold. For all $t\geq 0$, we have
    $
        \|\Delta\hat{\mathbf{e}}(t)\|\leq L_\xi L_\zeta\|\Delta\hat{\mathbf{x}}_0\|\text{exp}(L_f\|\mathbf{X}\|_{TV,[0,t]}).
    $
\end{corollary}
The above theoretical results imply that the Corrector is incrementally stable. Additionally, differences in initialization never blow up faster than exponentially. The stability of the full framework follows immediately since the Predictor cancels in differences, which is stated in the following result. We denote by $\mathbf x_{c,1}(t)=\hat{\mathbf x}(t)+\hat{\mathbf{e}}_1(t)$ and $\mathbf x_{c,2}(t)=\hat{\mathbf x}(t)+\hat{\mathbf{e}}_2(t)$ any two predictions. Hence, the stability of the corrected trajectories is described in the following.
\begin{corollary}\label{corollary_2}
    Let all assumptions hold. For any $t\geq 0$, it holds true that
    $
        \|\mathbf x_{c,1}(t)-\mathbf x_{c,2}(t)\|=\|\Delta\hat{\mathbf{e}}(t)\|\leq L_\xi L_\zeta\|\Delta\hat{\mathbf{x}}_0\|\text{exp}(L_f\|\mathbf{X}\|_{TV,[0,t]}).
    $
\end{corollary}
We now turn to the convergence of the error dynamics as stability alone does not imply convergence. We first introduce a standard contraction condition. 
\begin{assumption}\label{assumption_5}
    There exists $\lambda\geq 0$ such that
    $
        \langle\mathbf z_1-\mathbf z_2,f_\theta(\mathbf{z}_1)-f_\theta(\mathbf{z}_2)\rangle\leq -\lambda\|\mathbf z_1-\mathbf{z}_2\|^2, \forall \mathbf{z}_1,\mathbf{z}_2.
    $
\end{assumption}
The above assumption imposes a global strong monotonicity (one-sided contraction) condition on the latent vector field $f_\theta$, ensuring that latent trajectories are driven toward each other and enabling convergence. Without such a condition, one can only guarantee incremental stability (bounded growth), while convergence is generally impossible for nonlinear systems. This assumption is standard in contraction and incremental stability theory and is well aligned with the role of the Corrector, which is designed to damp residual prediction errors. Moreover, it can be satisfied or approximated by common Neural CDE parameterizations, such as linear damping terms or residual architectures with spectral normalization. 
We next present the exponential convergence of the latent Corrector. 
\begin{theorem}\label{theorem_2}
    Let all assumptions hold. Also, suppose that $\|\mathbf{X}\|_{TV,[0,\infty)}=\infty$. Then we have the following:
    $
        \|\Delta\mathbf z(t)\|\leq L_\zeta\|\Delta\hat{\mathbf{x}}_0\|\text{exp}(-\lambda\|\mathbf{X}\|_{TV,[0,t]}),
    $
    and
    $
        \text{lim}_{t\to\infty}\|\Delta\mathbf z(t)\|=0.
    $
\end{theorem}
We then similarly can get the convergence of the decoded error.
\begin{corollary}\label{corollary_3}
    Given the condition in Theorem~\ref{theorem_2}, we have the relationships:
    $
        \|\Delta\hat{\mathbf e}(t)\|\leq L_\xi L_\zeta\|\Delta\hat{\mathbf{x}}_0\|\text{exp}(-\lambda\|\mathbf{X}\|_{TV,[0,t]}),
    $
    and 
    $
        \text{lim}_{t\to\infty}\|\Delta\hat{\mathbf e}(t)\|=0.
    $
\end{corollary}
Hence, we can obtain the convergence of the full framework in the following statement. 
\begin{corollary}\label{corollary_4}
    Let all assumptions hold. For any $t\geq 0$, it holds true that
    $
        \text{lim}_{t\to\infty}\|\mathbf x_{c,1}(t)-\mathbf x_{c,2}(t)\|=0.
    $
\end{corollary}
From the above results, we can see that the convergence rate depends on $\|\mathbf{X}\|_{TV}$ rather than the physical time. Therefore, smoother predictor trajectories lead to faster and more robust convergence. Also, the encoder errors affect only initialization, whereas the decoder smoothness directly scales stability and convergence rates. Our theoretical results also justify the two regularization techniques we adopt in this work. The Neural CDE Corrector with encoder--decoder parameterization is incrementally stable under mild Lipschitz assumptions, guaranteeing bounded corrected trajectories. Moreover, if the latent vector field is contractive, the Corrector converges exponentially with respect to the total variation of the Predictor-induced control path, yielding a stable and convergent Predictor--Corrector forecasting framework.

\section{Results}

We now evaluate whether path-conditioned residual dynamics improve both interpolation and long-horizon extrapolation across diverse Predictors (Appendix~\ref{adx:pred_diversity}). The Neural CDE Corrector is paired with NODE \citep{chen2019neuralordinarydifferentialequations}, ContiFormer \citep{chen2024contiformercontinuoustimetransformerirregular}, and DLinear \citep{zeng2022transformerseffectivetimeseries} on synthetic, physics simulation, and LTSF datasets, respectively.


\subsection{Interpolation \& extrapolation} \label{subsec:inter_and_extra_polation}
Interpolation evaluates correction within the training horizon (first 50 steps); extrapolation reports the maximum timestep $t^*$ at which the Corrector still achieves $\geq 3\%$ MSE reduction over the Predictor.

\subsection{Synthetic datasets}\label{sec:synth_data}

\begin{table}[t]
\centering
\caption{%
  MSE without (w/o) and with (w/) corrector for \textsc{FHN} and \textsc{Glycolytic} (NODE predictor),
  combining interpolation ($0$--$50$) and extrapolation ($0$--$t^*$) regimes.
  w/ values show mean${\scriptstyle\pm\text{std}}$ over 3 runs; w/o is deterministic (single Predictor).
}
\label{tab:fhn_glycolytic}
\setlength{\tabcolsep}{2pt}
\renewcommand{\arraystretch}{1.05}
\scriptsize
\resizebox{\textwidth}{!}{\begin{tabular}{l|l|l|c|c|c|c!{\vrule width 1.5pt}c|c|c|c}
\toprule
\multirow{2}{*}{\textbf{Dataset}} & \multirow{2}{*}{\textbf{Corrector}} & & \multicolumn{4}{c|}{\textbf{Interpolation} ($0$--$50$)} & \multicolumn{4}{c}{\textbf{Extrapolation} ($0$--$t^*$)} \\
\cline{4-11}
& & & 20\% & 50\% & 80\% & 100\% & 20\% & 50\% & 80\% & 100\% \\
\midrule
\multirow{12}{*}{\textsc{FHN}}
 & \multirow{3}{*}{SLCDE}
   & w/o & 0.2257 & 0.1611 & 0.1508 & 0.1372 & 0.2465 & 0.1748 & 0.1884 & 0.1501 \\
 & & w/  & $0.2006{\scriptstyle\pm.012}$ & $0.1505{\scriptstyle\pm.008}$ & $0.1409{\scriptstyle\pm.005}$ & $0.1325{\scriptstyle\pm.004}$ & $0.2381{\scriptstyle\pm.017}$ & $0.1665{\scriptstyle\pm.010}$ & $0.1814{\scriptstyle\pm.009}$ & $0.1453{\scriptstyle\pm.006}$ \\
 & & $0$--$t|\%$ & $0$--$50|11\%$ & $0$--$50|7\%$ & $0$--$50|7\%$ & $0$--$50|3\%$ & $0$--$60|3\%$ & $0$--$55|5\%$ & $\mathbf{0}$--$\mathbf{150}|4\%$ & $0$--$55|3\%$ \\
\cmidrule{2-11}
 & \multirow{3}{*}{Latent ODE}
   & w/o & 0.2257 & 0.1611 & 0.1508 & 0.1372 & 0.2184 & 0.1470 & 0.1845 & 0.2102 \\
 & & w/  & $0.2227{\scriptstyle\pm.015}$ & $0.1644{\scriptstyle\pm.009}$ & $0.1534{\scriptstyle\pm.006}$ & $0.1401{\scriptstyle\pm.005}$ & $0.2115{\scriptstyle\pm.018}$ & $0.1417{\scriptstyle\pm.011}$ & $0.1783{\scriptstyle\pm.008}$ & $0.2014{\scriptstyle\pm.013}$ \\
 & & $0$--$t|\%$ & $0$--$50|1\%$ & $0$--$50|{-2}\%$ & $0$--$50|{-2}\%$ & $0$--$50|{-2}\%$ & $0$--$45|3\%$ & $0$--$40|4\%$ & $0$--$130|3\%$ & $0$--$5|4\%$ \\
\cmidrule{2-11}
 & \multirow{3}{*}{NRDE}
   & w/o & 0.2257 & 0.1611 & 0.1508 & 0.1372 & 0.2902 & 0.1829 & 0.1796 & 0.1500 \\
 & & w/  & $\mathbf{0.0947}{\scriptstyle\pm.005}$ & $\mathbf{0.0694}{\scriptstyle\pm.002}$ & $\mathbf{0.0670}{\scriptstyle\pm.003}$ & $\mathbf{0.0630}{\scriptstyle\pm.001}$ & $0.2741{\scriptstyle\pm.022}$ & $0.1739{\scriptstyle\pm.009}$ & $0.1714{\scriptstyle\pm.007}$ & $0.1263{\scriptstyle\pm.008}$ \\
 & & $0$--$t|\%$ & $0$--$50|58\%$ & $0$--$50|57\%$ & $0$--$50|56\%$ & $0$--$50|54\%$ & $\mathbf{0}$--$\mathbf{140}|6\%$ & $0$--$100|5\%$ & $0$--$85|5\%$ & $0$--$70|16\%$ \\
\cmidrule{2-11}
 & \multirow{3}{*}{Neural CDE}
   & w/o & 0.2257 & 0.1611 & 0.1508 & 0.1372 & 0.2420 & 0.1779 & 0.1709 & 0.1667 \\
 & & w/  & $0.1647{\scriptstyle\pm.010}$ & $0.1289{\scriptstyle\pm.007}$ & $0.1041{\scriptstyle\pm.003}$ & $0.0979{\scriptstyle\pm.004}$ & $0.2309{\scriptstyle\pm.019}$ & $0.1722{\scriptstyle\pm.006}$ & $0.1624{\scriptstyle\pm.011}$ & $0.1585{\scriptstyle\pm.005}$ \\
 & & $0$--$t|\%$ & $0$--$50|27\%$ & $0$--$50|20\%$ & $0$--$50|31\%$ & $0$--$50|29\%$ & $0$--$75|5\%$ & $\mathbf{0}$--$\mathbf{155}|3\%$ & $0$--$75|5\%$ & $\mathbf{0}$--$\mathbf{150}|5\%$ \\
\midrule
\multirow{12}{*}{\textsc{Glycolytic}}
 & \multirow{3}{*}{SLCDE}
   & w/o & 0.0131 & 0.0100 & 0.0098 & 0.0098 & 0.0353 & 0.0177 & 0.0116 & 0.0115 \\
 & & w/  & $0.0094{\scriptstyle\pm.0007}$ & $0.0098{\scriptstyle\pm.0004}$ & $0.0080{\scriptstyle\pm.0005}$ & $0.0080{\scriptstyle\pm.0002}$ & $0.0338{\scriptstyle\pm.0021}$ & $0.0171{\scriptstyle\pm.0008}$ & $0.0111{\scriptstyle\pm.0006}$ & $0.0111{\scriptstyle\pm.0004}$ \\
 & & $0$--$t|\%$ & $0$--$50|29\%$ & $0$--$50|1\%$ & $0$--$50|19\%$ & $0$--$50|18\%$ & $\mathbf{0}$--$\mathbf{195}|4\%$ & $0$--$15|3\%$ & $0$--$95|4\%$ & $0$--$95|4\%$ \\
\cmidrule{2-11}
 & \multirow{3}{*}{Latent ODE}
   & w/o & 0.0131 & 0.0100 & 0.0098 & 0.0098 & 0.0213 & 0.0177 & 0.0352 & 0.0102 \\
 & & w/  & $0.0077{\scriptstyle\pm.0006}$ & $0.0098{\scriptstyle\pm.0003}$ & $0.0075{\scriptstyle\pm.0003}$ & $0.0078{\scriptstyle\pm.0002}$ & $0.0201{\scriptstyle\pm.0014}$ & $0.0171{\scriptstyle\pm.0009}$ & $0.0341{\scriptstyle\pm.0027}$ & $0.0097{\scriptstyle\pm.0005}$ \\
 & & $0$--$t|\%$ & $0$--$50|41\%$ & $0$--$50|1\%$ & $0$--$50|24\%$ & $0$--$50|20\%$ & $0$--$110|6\%$ & $0$--$15|3\%$ & $0$--$145|3\%$ & $0$--$85|5\%$ \\
\cmidrule{2-11}
 & \multirow{3}{*}{NRDE}
   & w/o & 0.0131 & 0.0100 & 0.0098 & 0.0098 & 0.0356 & 0.0180 & 0.0356 & 0.0347 \\
 & & w/  & $\mathbf{0.0075}{\scriptstyle\pm.0004}$ & $\mathbf{0.0064}{\scriptstyle\pm.0002}$ & $\mathbf{0.0062}{\scriptstyle\pm.0002}$ & $\mathbf{0.0063}{\scriptstyle\pm.0001}$ & $0.0345{\scriptstyle\pm.0026}$ & $0.0170{\scriptstyle\pm.0010}$ & $0.0344{\scriptstyle\pm.0022}$ & $0.0334{\scriptstyle\pm.0019}$ \\
 & & $0$--$t|\%$ & $0$--$50|43\%$ & $0$--$50|35\%$ & $0$--$50|37\%$ & $0$--$50|36\%$ & $0$--$155|3\%$ & $0$--$110|5\%$ & $\mathbf{0}$--$\mathbf{150}|3\%$ & $0$--$115|4\%$ \\
\cmidrule{2-11}
 & \multirow{3}{*}{Neural CDE}
   & w/o & 0.0131 & 0.0100 & 0.0098 & 0.0098 & 0.0159 & 0.0204 & 0.0135 & 0.0160 \\
 & & w/  & $0.0079{\scriptstyle\pm.0006}$ & $0.0073{\scriptstyle\pm.0002}$ & $0.0072{\scriptstyle\pm.0003}$ & $0.0068{\scriptstyle\pm.0001}$ & $0.0146{\scriptstyle\pm.0011}$ & $0.0197{\scriptstyle\pm.0013}$ & $0.0128{\scriptstyle\pm.0007}$ & $0.0154{\scriptstyle\pm.0009}$ \\
 & & $0$--$t|\%$ & $0$--$50|40\%$ & $0$--$50|27\%$ & $0$--$50|26\%$ & $0$--$50|30\%$ & $0$--$100|8\%$ & $\mathbf{0}$--$\mathbf{115}|3\%$ & $0$--$100|5\%$ & $\mathbf{0}$--$\mathbf{125}|4\%$ \\
\bottomrule
\end{tabular}}
\end{table}

We evaluate four continuous-time Corrector architectures on synthetic data generated from four multivariate ODE systems: \textsc{Lorenz}, \textsc{LVolt}, \textsc{FHN}, and \textsc{Glycolytic}. Table~\ref{tab:fhn_glycolytic} reports \textsc{FHN} and \textsc{Glycolytic}, while Appendix~\ref{adx:lorenz_lvolt} reports \textsc{Lorenz} and \textsc{LVolt}. NODE \citep{chen2019neuralordinarydifferentialequations} is used as the Predictor and trained on the first 50 timesteps. Each Corrector is trained on the first 50 timesteps (interpolation regime) on the same splits; all four levels of sparsity from Section~\ref{sec:sparse_ctrl} are applied during training. Dataset details are given in Table~\ref{tab:data_detail} and ODE equations in Appendix~\ref{adx:synth_data}. Hyperparameters $\kappa$ and $\eta$ for all settings are in Table~\ref{tab:synthetic_hyperparam}. Further training details are in Appendices~\ref{adx:node_train} and~\ref{adx:neural_cde}.

\begin{wrapfigure}[16]{r}{0.38\textwidth}
    \vspace{-14pt}
    \centering
    \includegraphics[width=0.95\linewidth]{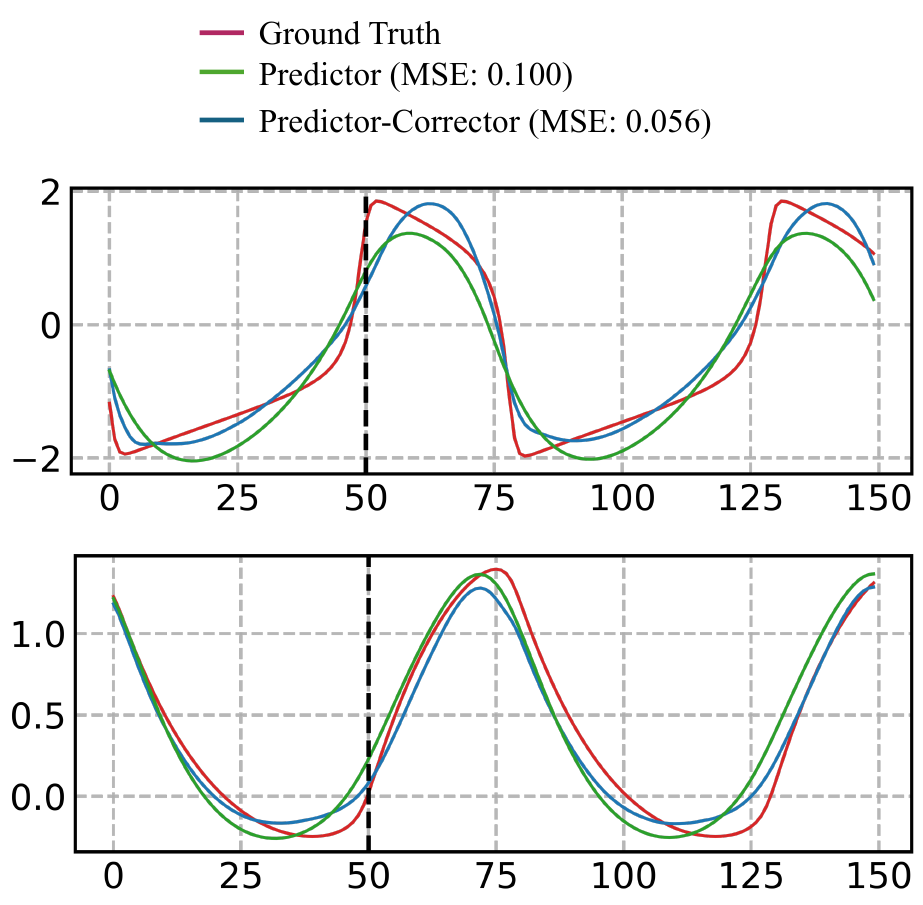}
    \caption{NODE on an \textsc{FHN} test trajectory with and without the Corrector.}
    \label{fig:fhn_perform}
\end{wrapfigure}

\paragraph{Corrector baselines.}
We compare our Neural CDE Corrector against three architectures that differ in how they use the predictor trajectory $\hat{\mathbf{x}}_{0:T}$.
A high-level comparison of these continuous-time baselines with Neural CDE is given in Appendix~\ref{adx:ct_baselines}.
\textbf{Latent ODE} \citep{rubanova2019latentodesirregularlysampledtime} encodes $\hat{\mathbf{x}}_{0:T}$ into an initial latent state via an ODE-RNN and evolves it with NODE ($\dot{z}=f_\theta(z)$) without explicit path-conditioning; the forecast trajectory no longer influences the residual dynamics after encoding.
\textbf{Structured Linear CDE (SLCDE)} \citep{walker2025structured} uses a linear controlled differential equation where the vector field is a linear function of the control path, providing path-driven dynamics under structural constraints.
\textbf{Neural RDE (NRDE)} \citep{morrill2021neural} uses depth-2 log-signatures of the control path to capture higher-order path interactions. At the finest partition, these features reduce to per-step forecast increments, the same local information Neural CDE exploits continuously, making NRDE a strong but structurally closest baseline.
All four Correctors share the same decoder, hidden dimension, data pipeline, loss, and optimiser; only the encoder/dynamics module differs.

\paragraph{Results.}
Table~\ref{tab:fhn_glycolytic} reports \textsc{FHN} and \textsc{Glycolytic}; \textsc{Lorenz} and \textsc{LVolt} are in Appendix~\ref{adx:lorenz_lvolt}.
Without explicit path-conditioning, the Latent ODE Corrector is unreliable, yielding negligible or negative gains on \textsc{FHN} ($\leq 1\%$) and increasing NODE's MSE on \textsc{LVolt} by 95--429\% (Appendix~\ref{adx:lorenz_lvolt}).
In contrast, the path-driven Correctors, SLCDE, NRDE, and Neural CDE, consistently reduce NODE's interpolation error on all four systems, showing that the forecast path contains useful information about the residual dynamics.
This supports the main modeling premise that residual dynamics are trajectory-dependent and cannot be captured without explicit conditioning on the Predictor forecast.

Among path-driven models, NRDE is the strongest interpolation baseline on several systems, but Neural CDE provides the best interpolation--extrapolation balance: it gives substantial interpolation reductions across all four systems while reaching the longest extrapolation horizons in key \textsc{Lorenz} and \textsc{FHN} settings. Fig.~\ref{fig:fhn_perform} illustrates per-trajectory correction; stress tests confirming bounded error growth are in Appendix~\ref{adx:stress_test}.

\subsection{Physics simulation}\label{sec:phy_data}

\begin{table}[t]
\centering
\caption{%
  MSE without (w/o) and with (w/) corrector for \textsc{Walker2D} and \textsc{Pen},
  combining interpolation ($0$--$50$) and extrapolation ($0$--$t^*$) regimes.
  $t^*$ is the maximum timestep sustaining $\geq 3\%$ cumulative MSE reduction.
  w/ values show mean${\scriptstyle\pm\text{std}}$ over 3 runs; w/o is deterministic (single Predictor).
  Best w/ per column in \textbf{bold} (interpolation); longest $t^*$ per column in
  \textbf{bold} (extrapolation).
}
\label{tab:walker2d_pen}
\setlength{\tabcolsep}{1.5pt}
\renewcommand{\arraystretch}{1.05}
\scriptsize
\resizebox{\textwidth}{!}{\begin{tabular}{l|l|l|c|c|c|c!{\vrule width 1.5pt}c|c|c|c}
\toprule
\multirow{2}{*}{\textbf{Dataset}} & \multirow{2}{*}{\textbf{Corrector}} & & \multicolumn{4}{c|}{\textbf{Interpolation} ($0$--$50$)} & \multicolumn{4}{c}{\textbf{Extrapolation} ($0$--$t^*$)} \\
\cline{4-11}
& & & 20\% & 50\% & 80\% & 100\% & 20\% & 50\% & 80\% & 100\% \\
\midrule
\multirow{12}{*}{\textsc{Walker2D}}
 & \multirow{3}{*}{SLCDE}
   & w/o & 1.827 & 0.573 & 0.408 & 0.165 & 2.467 & 0.897 & 0.470 & 0.275 \\
 & & w/ & $0.856{\scriptstyle\pm.071}$ & $0.399{\scriptstyle\pm.024}$ & $0.261{\scriptstyle\pm.018}$ & $0.139{\scriptstyle\pm.007}$ & $2.369{\scriptstyle\pm.189}$ & $0.868{\scriptstyle\pm.052}$ & $0.451{\scriptstyle\pm.031}$ & $0.265{\scriptstyle\pm.009}$ \\
 & & $0$--$t^*|\%$ & $0$--$50|53\%$ & $0$--$50|30\%$ & $0$--$50|36\%$ & $0$--$50|16\%$ & $0$--$150|4\%$ & $0$--$125|3\%$ & $0$--$115|4\%$ & $0$--$85|4\%$ \\
\cmidrule{2-11}
 & \multirow{3}{*}{Latent ODE}
   & w/o & 1.827 & 0.573 & 0.408 & 0.165 & 1.774 & 0.665 & 0.450 & 0.275 \\
 & & w/ & $\mathbf{0.211}{\scriptstyle\pm.017}$ & $0.288{\scriptstyle\pm.023}$ & $0.161{\scriptstyle\pm.009}$ & $0.075{\scriptstyle\pm.003}$ & $1.718{\scriptstyle\pm.137}$ & $0.645{\scriptstyle\pm.038}$ & $0.431{\scriptstyle\pm.024}$ & $0.256{\scriptstyle\pm.011}$ \\
 & & $0$--$t^*|\%$ & $0$--$50|88\%$ & $0$--$50|50\%$ & $0$--$50|61\%$ & $0$--$50|54\%$ & $0$--$125|3\%$ & $0$--$90|3\%$ & $0$--$110|4\%$ & $0$--$85|7\%$ \\
\cmidrule{2-11}
 & \multirow{3}{*}{NRDE}
   & w/o & 1.827 & 0.573 & 0.408 & 0.165 & 1.868 & 0.577 & 0.451 & 0.245 \\
 & & w/ & $0.234{\scriptstyle\pm.019}$ & $\mathbf{0.123}{\scriptstyle\pm.007}$ & $\mathbf{0.086}{\scriptstyle\pm.004}$ & $\mathbf{0.047}{\scriptstyle\pm.002}$ & $1.806{\scriptstyle\pm.144}$ & $\mathbf{0.408}{\scriptstyle\pm.029}$ & $\mathbf{0.371}{\scriptstyle\pm.021}$ & $\mathbf{0.233}{\scriptstyle\pm.008}$ \\
 & & $0$--$t^*|\%$ & $0$--$50|87\%$ & $0$--$50|79\%$ & $0$--$50|79\%$ & $0$--$50|72\%$ & $0$--$130|3\%$ & $0$--$70|29\%$ & $0$--$75|18\%$ & $0$--$70|5\%$ \\
\cmidrule{2-11}
 & \multirow{3}{*}{Neural CDE}
   & w/o & 1.827 & 0.573 & 0.408 & 0.165 & 2.890 & 1.700 & 0.778 & 0.869 \\
 & & w/ & $0.721{\scriptstyle\pm.058}$ & $0.285{\scriptstyle\pm.019}$ & $0.149{\scriptstyle\pm.008}$ & $0.053{\scriptstyle\pm.003}$ & $2.680{\scriptstyle\pm.214}$ & $1.630{\scriptstyle\pm.098}$ & $0.750{\scriptstyle\pm.042}$ & $0.838{\scriptstyle\pm.056}$ \\
 & & $0$--$t^*|\%$ & $0$--$50|61\%$ & $0$--$50|50\%$ & $0$--$50|64\%$ & $0$--$50|68\%$ & $\mathbf{0}$--$\mathbf{190}|7\%$ & $\mathbf{0}$--$\mathbf{180}|4\%$ & $\mathbf{0}$--$\mathbf{130}|4\%$ & $\mathbf{0}$--$\mathbf{140}|4\%$ \\
\midrule
\multirow{12}{*}{\textsc{Pen}}
 & \multirow{3}{*}{SLCDE}
   & w/o & 0.349 & 0.142 & 0.128 & 0.124 & 0.213 & 0.102 & 0.103 & 0.099 \\
 & & w/ & $0.147{\scriptstyle\pm.011}$ & $0.105{\scriptstyle\pm.005}$ & $0.093{\scriptstyle\pm.004}$ & $0.088{\scriptstyle\pm.003}$ & $0.205{\scriptstyle\pm.016}$ & $0.094{\scriptstyle\pm.005}$ & $0.093{\scriptstyle\pm.006}$ & $0.090{\scriptstyle\pm.003}$ \\
 & & $0$--$t^*|\%$ & $0$--$50|58\%$ & $0$--$50|26\%$ & $0$--$50|28\%$ & $0$--$50|29\%$ & $0$--$90|4\%$ & $0$--$80|8\%$ & $0$--$70|10\%$ & $0$--$70|10\%$ \\
\cmidrule{2-11}
 & \multirow{3}{*}{Latent ODE}
   & w/o & 0.349 & 0.142 & 0.128 & 0.124 & 0.182 & 0.081 & 0.083 & 0.078 \\
 & & w/ & $0.101{\scriptstyle\pm.008}$ & $0.078{\scriptstyle\pm.004}$ & $0.074{\scriptstyle\pm.003}$ & $0.072{\scriptstyle\pm.002}$ & $0.159{\scriptstyle\pm.013}$ & $0.076{\scriptstyle\pm.004}$ & $0.076{\scriptstyle\pm.005}$ & $0.071{\scriptstyle\pm.003}$ \\
 & & $0$--$t^*|\%$ & $0$--$50|71\%$ & $0$--$50|45\%$ & $0$--$50|43\%$ & $0$--$50|42\%$ & $0$--$110|13\%$ & $0$--$120|6\%$ & $0$--$95|8\%$ & $0$--$100|9\%$ \\
\cmidrule{2-11}
 & \multirow{3}{*}{NRDE}
   & w/o & 0.349 & 0.142 & 0.128 & 0.124 & 0.130 & 0.069 & 0.070 & 0.071 \\
 & & w/ & $\mathbf{0.081}{\scriptstyle\pm.006}$ & $\mathbf{0.060}{\scriptstyle\pm.002}$ & $0.055{\scriptstyle\pm.003}$ & $0.057{\scriptstyle\pm.002}$ & $0.126{\scriptstyle\pm.009}$ & $0.066{\scriptstyle\pm.004}$ & $0.067{\scriptstyle\pm.003}$ & $0.068{\scriptstyle\pm.004}$ \\
 & & $0$--$t^*|\%$ & $0$--$50|77\%$ & $0$--$50|58\%$ & $0$--$50|57\%$ & $0$--$50|54\%$ & $0$--$175|3\%$ & $0$--$165|4\%$ & $0$--$140|5\%$ & $0$--$130|4\%$ \\
\cmidrule{2-11}
 & \multirow{3}{*}{Neural CDE}
   & w/o & 0.349 & 0.142 & 0.128 & 0.124 & 0.117 & 0.061 & 0.057 & 0.056 \\
 & & w/ & $0.150{\scriptstyle\pm.012}$ & $0.072{\scriptstyle\pm.004}$ & $\mathbf{0.050}{\scriptstyle\pm.002}$ & $\mathbf{0.055}{\scriptstyle\pm.003}$ & $0.110{\scriptstyle\pm.008}$ & $0.059{\scriptstyle\pm.003}$ & $0.054{\scriptstyle\pm.004}$ & $0.054{\scriptstyle\pm.002}$ \\
 & & $0$--$t^*|\%$ & $0$--$50|57\%$ & $0$--$50|49\%$ & $0$--$50|61\%$ & $0$--$50|55\%$ & $\mathbf{0}$--$\mathbf{250}|6\%$ & $\mathbf{0}$--$\mathbf{190}|3\%$ & $\mathbf{0}$--$\mathbf{205}|4\%$ & $\mathbf{0}$--$\mathbf{220}|4\%$ \\
\bottomrule
\end{tabular}}
\end{table}

We next test whether the same path-conditioned correction transfers from low-dimensional ODE systems to higher-dimensional simulated control trajectories. We evaluate the four Corrector architectures from Section~\ref{sec:synth_data} on four MuJoCo environments \citep{todorov2012mujoco,towers2024gymnasiumstandardinterfacereinforcement}: \textsc{Walker2D} (17D), \textsc{Pen} (45D), \textsc{Hopper} (11D), and \textsc{Hammer} (46D). Appendix~\ref{adx:ct_baselines} summarizes how these continuous-time Corrector baselines differ in their conditioning on the Predictor forecast path. ContiFormer \citep{chen2024contiformercontinuoustimetransformerirregular}, a continuous-time transformer, is used as the Predictor and trained to predict 100 timesteps. Each dataset consists of 2,000 trajectories of 300 regularly sampled timesteps. We apply an 80/20 train/test split; all four sparsity levels from Section~\ref{sec:sparse_ctrl} are simulated during training. Each Corrector is trained on the first 50 timesteps (interpolation regime) on the same splits using the shared decoder, hidden dimension, and training configuration described in Section~\ref{sec:synth_data}. Data generation and training details are in Appendices~\ref{adx:phy_sim},~\ref{adx:conti_train}, and~\ref{adx:neural_cde}.

\paragraph{Results.}
Table~\ref{tab:walker2d_pen} reports \textsc{Walker2D} and \textsc{Pen} results; \textsc{Hopper} and \textsc{Hammer} are in Appendix~\ref{adx:hopper_hammer}.
Across these simulated control tasks, NRDE often provides the strongest interpolation, while Neural CDE is more reliable for long-horizon correction.
This setting is deliberately harder than the synthetic systems: state dimensions are larger, and forecast errors can couple across coordinates. The recurring advantage of path-driven correction therefore suggests that continuous conditioning is not an artifact of low-dimensional ODE benchmarks.
On high-dimensional \textsc{Pen}, for example, Neural CDE sustains improvement substantially farther than NRDE, and this extrapolation advantage persists across sparsity levels.
Appendix~\ref{adx:hopper_hammer} shows the same interpolation--extrapolation tradeoff on \textsc{Hopper} and \textsc{Hammer}; qualitative \textsc{Pen} trajectories are in Appendix~\ref{adx:pen_viz}. The hyperparameters $\eta$ and $\kappa$ used for each setting in Table~\ref{tab:walker2d_pen} are reported in Table~\ref{tab:mujoco_hyperparam}.

\begin{figure}[ht]
    \centering
    \includegraphics[width=0.85\columnwidth]{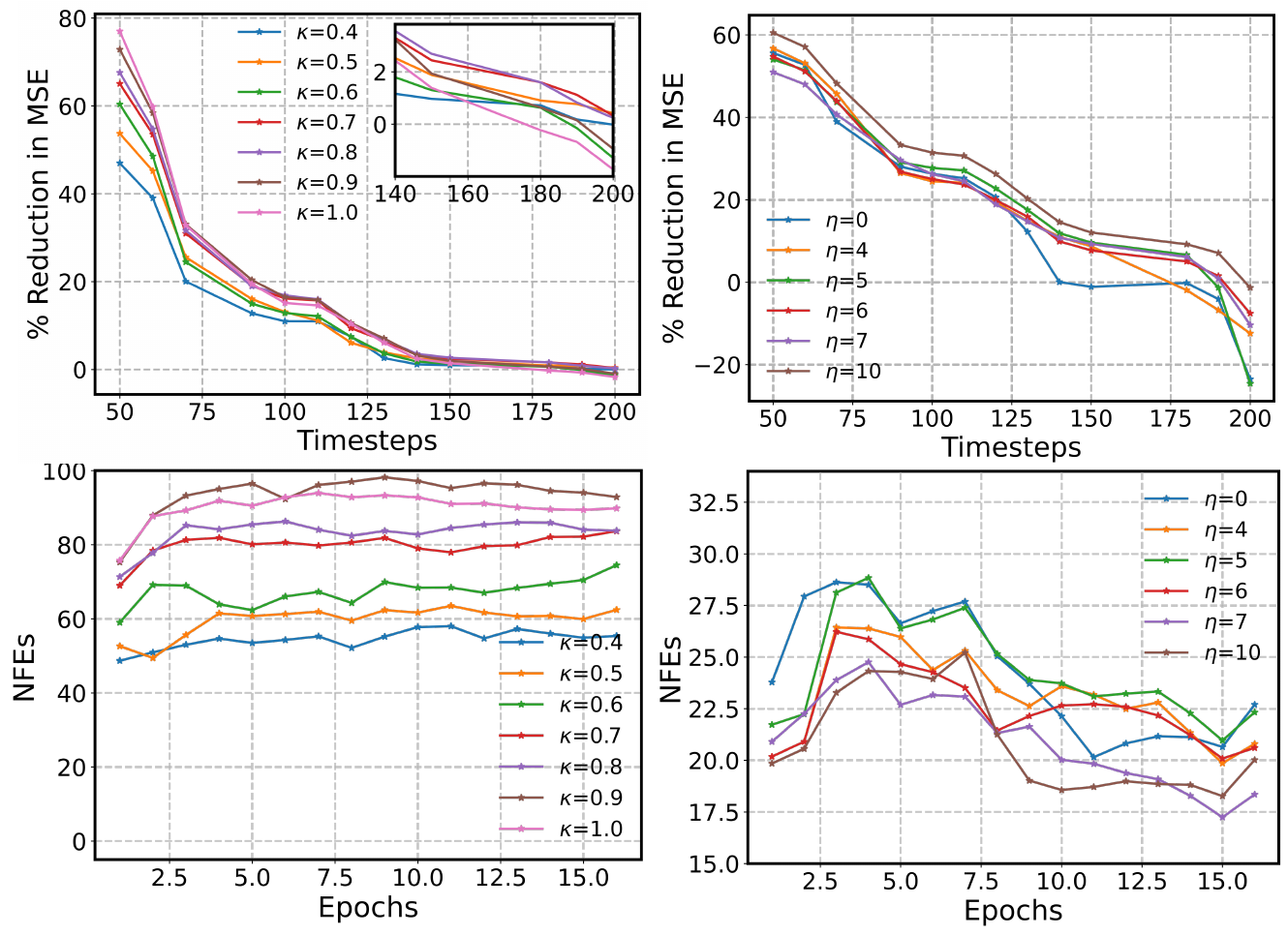}
\caption{Control-path regularization on \textsc{Walker2D}. \textbf{(a) Upper left:} MSE reduction (\%) vs.\ timestep for varying $\kappa$, $\eta{=}0$. \textbf{(b) Lower left:} NFEs vs.\ epoch for varying $\kappa$, $\eta{=}0$. \textbf{(c) Upper right:} MSE reduction (\%) vs.\ timestep for varying $\eta$, $\kappa{=}1.0$. \textbf{(d) Lower right:} NFEs vs.\ epoch for varying $\eta$, $\kappa{=}1.0$. Both strategies extend the extrapolation horizon and reduce NFEs at the cost of interpolation fidelity.}
    \label{fig:nfe_reduction}
\end{figure}

\paragraph{Sparse control paths.}
 To isolate the effect of Sparse Control Paths, we study \textsc{Walker2D} with 100\% observed time points. With $\kappa=0.7$ and $\eta=0$, the Corrector reduces MSE by 68\% over timesteps 0--50 and by 4\% over 0--140 (Table~\ref{tab:walker2d_pen}). Figure~\ref{fig:nfe_reduction}(a) shows the tradeoff: no regularization ($\kappa=1.0$) gives stronger interpolation (78\% over 0--50) but fails beyond 175 timesteps, whereas $\kappa=0.7$ maintains positive correction to 200 timesteps. Figure~\ref{fig:nfe_reduction}(b) shows the efficiency benefit: smaller $\kappa$ increases path sparsity and reduces NFEs, with $\kappa=0.7$ stabilizing near 80 NFEs after early epochs compared with over 90 for $\kappa=1.0$. Thus, Sparse Control Paths improve extrapolation and reduce computation, but $\kappa$ remains a tuned hyperparameter.

\paragraph{Variable-length control paths.}
For \textsc{Walker2D} at 20\% observations ($\kappa=1.0$, $\eta=10$), Fig.~\ref{fig:nfe_reduction}(c) shows that $\eta=0$ yields poor extrapolation (drops below zero near $t=140$), while $\eta=10$ maintains positive correction to $t=200$. Like $\kappa$, $\eta$ must be tuned. Fig.~\ref{fig:nfe_reduction}(d) shows that larger $\eta$ reduces NFEs by exposing the integrator to shorter paths on average, yielding faster training alongside improved extrapolation.

Sparse control paths are most effective at high observation rates (50–100\%), whereas variable-length paths improve sparsely observed regimes. Appendix~\ref{adx:pareto_curves} quantifies the NFE--extrapolation trade-off via Pareto curves and Appendix~\ref{adx:clock_time} confirms the wall-clock speedup.

\subsection{Time series forecasting}\label{sec:fore_data}

\begin{table*}[ht]
\centering
\caption{Multivariate long-term forecasting errors (MSE/MAE; lower is better) for five LTSF datasets, averaged over all forecast horizons per dataset. All Correctors use the same frozen DLinear Predictor and are parameter-matched ($\approx$530\,K). Best value per row is highlighted in bold. Per-horizon results are in Appendix Table~\ref{tab:corrector_full}.}
\vspace{0.15cm}
\resizebox{\textwidth}{!}{%
\begin{tabular}{c|cc|cc|cc|cc|cc}
\toprule
Dataset
& \multicolumn{2}{c|}{DLinear (w/o)}
& \multicolumn{2}{c|}{+MLP}
& \multicolumn{2}{c|}{+GRU}
& \multicolumn{2}{c|}{+Transformer}
& \multicolumn{2}{c}{+Neural CDE} \\
\midrule
Metric & MSE & MAE & MSE & MAE & MSE & MAE & MSE & MAE & MSE & MAE \\
\midrule
Exchange & 0.370 & 0.419 & 0.944 & 0.709 & 0.646 & 0.619 & 0.549 & 0.536 & \textbf{0.242} & \textbf{0.355} \\
\midrule
ETTm2    & 0.284 & 0.346 & 1.090 & 0.672 & 0.985 & 0.610 & 1.106 & 0.712 & \textbf{0.271} & \textbf{0.338} \\
\midrule
ETTh2    & 0.470 & 0.465 & 3.324 & 1.363 & 2.182 & 1.091 & 2.339 & 1.228 & \textbf{0.458} & \textbf{0.458} \\
\midrule
ILI      & 2.403 & 1.110 & 2.548 & 1.095 & \textbf{2.298} & \textbf{1.062} & 2.450 & 1.109 & 2.350 & 1.080 \\
\midrule
Weather  & 0.248 & 0.301 & 0.245 & 0.307 & 0.240 & 0.301 & 0.253 & 0.323 & \textbf{0.238} & \textbf{0.286} \\
\bottomrule
\end{tabular}%
}
\label{tab:forecast_results}
\end{table*}

Finally, we evaluate whether the same path-conditioned Corrector can improve a purely discrete-time Predictor. We apply the Corrector to frozen DLinear \citep{zeng2022transformerseffectivetimeseries} forecasts on five LTSF datasets (Appendix~\ref{adx:forecast_dataset}): Exchange, ETTm2, ETTh2, ILI, and Weather. For ILI, forecast horizons are $T \in \{24,36,48,60\}$; for other datasets, $T \in \{96,192,336,720\}$ with the Corrector trained on the first $\{50,100,150,300\}$ timesteps (Appendix~\ref{adx:forecast_corr_hor}). Hyperparameters $\kappa$ and $\eta$ are in Tables~\ref{tab:forecast_hyperparam_1} and~\ref{tab:forecast_hyperparam_2}.

\paragraph{Results.} Table~\ref{tab:forecast_results} compares four parameter-matched Correctors ($\approx$530\,K; Appendix~\ref{adx:discrete_baselines}) against frozen DLinear, averaged over all horizons.
Neural CDE is the only Corrector that consistently improves DLinear across all five datasets, including the largest MSE reduction on Exchange and the only joint MSE/MAE reductions on ETTm2 and ETTh2; GRU is strongest only on ILI.
The MLP and Transformer Correctors degrade DLinear performance on all five datasets despite matching the Neural CDE parameter budget, indicating that continuous path conditioning provides more reliable residual correction than discrete alternatives.
The discrete baselines receive the same forecast sequence, but operate on sampled forecast values rather than on an interpolated path. Neural CDE instead conditions on the continuous forecast path, so its residual dynamics can respond to local increments and slopes between sampled time points.
Full per-horizon results are in Appendix Table~\ref{tab:corrector_full}; transformer baseline comparisons in Appendix~\ref{adx:full_forecast_baselines}; FEDformer results in Appendix~\ref{adx:ncde_fedformer}.
\paragraph{Discrete versus continuous correction.}
Among the discrete Corrector baselines, GRU is the closest analogue to Neural CDE because it also carries a hidden state forward while conditioning on the Predictor forecast sequence.
However, GRU updates its hidden state $h_i$ as $h_i=\mathrm{GRU}(h_{i-1},\hat{\mathbf{x}}_i)$, whereas Neural CDE evolves its hidden state $z_t$ according to $dz_t=f_\theta(z_t)dX_t$, equivalently $\frac{dz_t}{dt}=f_\theta(z_t)\frac{dX_t}{dt}$.
Thus, Neural CDE is driven by the local slope of the interpolated forecast path, while GRU and Transformer rely on the discrete forecast sequence unless time-gap or derivative features are explicitly added.
The same distinction also applies to the Transformer Corrector.
Although it can attend globally over the forecast horizon, residual correction may depend more on local error evolution, such as gradual drift, phase mismatch, and trend changes, than on global sequence mixing.
Together, these comparisons suggest that Neural CDE's advantage comes from its path-evolution inductive bias, not merely from conditioning on the same forecast sequence.



\section{Conclusion}

We introduced a Predictor-Corrector framework where a Neural CDE Corrector models forecast residuals as path-conditioned dynamics rather than unstructured noise. The key design choice is to keep the residual model continuously coupled to the Predictor forecast path, so correction can depend on how the forecast trajectory evolves over the horizon. This differs from autonomous latent correction, where the trajectory influences the hidden state only through an initial encoding, and from discrete Correctors that update only at sampled forecast values. Because the Corrector is driven by the forecast path, it can be attached to continuous- or discrete-time Predictors without modifying their architectures. Two control-path regularization strategies improve extrapolation and reduce training cost. Supported by stability and convergence guarantees, the framework consistently improves NODE and ContiFormer on synthetic and physics simulation datasets, and DLinear on real-world LTSF benchmarks, demonstrating predictor-agnostic applicability. Empirically, the advantage appears when correction must remain coupled to the forecast trajectory beyond the training horizon.

\section{Limitations}

The framework adds a second training stage after the Predictor is learned, and Neural CDE inference requires solving a controlled differential equation. Its benefit also depends on residual errors being structured and related to the forecast path; when Predictor errors are weakly path-dependent, correction may be limited. Finally, although the experiments cover synthetic dynamics, physics simulation, and LTSF benchmarks, they do not exhaust all forecasting domains, data scales, or deployment constraints.

\bibliographystyle{plainnat}
\bibliography{example_paper}

\newpage
\appendix
\onecolumn
\section{Auxiliary technical lemmas and missing proof}
In this section, we present the auxiliary technical lemmas that characterize the analysis of our main result. Additionally, we also give the complete proof for all lemmas and main results of the stability and convergence.
\begin{lemma}\label{lemma_1}
    The latent difference satisfies 
    \begin{equation}\label{eq_13}
        \Delta\mathbf z(t)=\Delta\mathbf z(t_0) + \int^{t}_{t_0}(f_\theta(\mathbf z_1(s))-f_\theta(\mathbf z_2(s)))d\mathbf{X}(s).
    \end{equation}
\end{lemma}
\begin{proof}
    Subtracting the two neural CDEs in Eq.~\ref{eq_5} initialized at $\mathbf z_1(t_0)$ and $\mathbf z_2(t_0)$ completes the proof.
\end{proof}
\begin{lemma}\label{lemma_2}
    Let Assumption~\ref{assumption_3} hold. For all $t\geq t_0\geq 0$, we have
    \begin{equation}
        \|\Delta\mathbf z(t)\|\leq \|\Delta\mathbf z(t_0)\| + \int^{t}_{t_0}L_g\|\Delta\mathbf z(s)\|d\|\mathbf{X}\|(s).
    \end{equation}
\end{lemma}
\begin{proof}
    Taking norms in Eq.~\ref{eq_13} and applying Assumption~\ref{assumption_3} leads to the result.
\end{proof}
We next present the proof for Theorem~\ref{theorem_1}.
\begin{proof}
    Define the scalar function $u(t):=\|\Delta\mathbf z(t)\|$. Recalling the conclusion from Lemma~\ref{lemma_2}, it becomes
    \begin{equation}\label{eq_16}
        u(t)\leq u(t_0)+\int^t_{t_0}L_fu(s)d\mu(s),
    \end{equation}
    where we define the nondecreasing function $\mu(t):=\|\mathbf{X}\|_{TV,[0,t]}$. Note that $\mu(0)=0$ and $\mu(t)$ is finite and nondecreasing. We apply the Gr\"onwall inequality, which states that if a nonnegative function $u(t)$ satisfies $u(t)\leq a+\int^t_{t_0}b u(s)d\mu(s)$, with constants $a,b\geq 0$ and nondecreasing $\mu$, then 
    \begin{equation}\label{eq_17}
        u(t)\leq a\text{exp}(b\mu(t)). 
    \end{equation}
    In our case, $a=u(t_0)=\|\Delta\mathbf z(t_0)\|$, $b=L_f$, and $\mu(t)=\|\mathbf{X}\|_{TV,[0,t]}$. Applying Eq.~\ref{eq_17} to Eq.~\ref{eq_16}, we obtain
    \begin{equation}
        u(t)\leq \|\Delta\mathbf z(t_0)\|\text{exp} (L_f\|\mathbf{X}\|_{TV,[0,t]}).
    \end{equation}
    Recalling $u(t)=\|\Delta\mathbf z(t)\|$ and using the fact that $\|\Delta\mathbf{z}(t_0)\|=\|\zeta_\phi(\hat{\mathbf{x}}_0)-\zeta_\phi(\hat{\mathbf{x}}'_0)\|$ with Assumption~\ref{assumption_2}, we get the desirable result.
\end{proof}
We next provide the proof for Corollary~\ref{corollary_1} and Corollary~\ref{corollary_2}.
\begin{proof}
    Applying Assumption~\ref{assumption_4} to the conclusion from Theorem~\ref{theorem_1} yields the conclusion.
\end{proof}
\begin{proof}
    The Predictor cancels exactly. We apply Corollary~\ref{corollary_1} to complete the proof.
\end{proof}
\begin{lemma}\label{lemma_3}
    Let Assumption~\ref{assumption_5} hold. The squared latent difference satisfies
    \begin{equation}\label{eq_22}
        d\|\Delta\mathbf z(t)\|^2\leq -2\lambda\|\Delta\mathbf z(t)\|^2d\|\mathbf{X}\|(t).
    \end{equation}
\end{lemma}
\begin{proof}
    From Lemma~\ref{lemma_1}, we obtain that 
    \begin{equation}
        d\Delta\mathbf z(t) = (f_\theta(\mathbf z_1)-f_\theta(\mathbf z_2))d\mathbf X.
    \end{equation}
    Then
    \begin{equation}
        d\|\Delta\mathbf z(t)\|^2=2\langle\Delta\mathbf{z}, f_\theta(\mathbf z_1)-f_\theta(\mathbf z_2\rangle d\mathbf{X}.
    \end{equation}
    Applying Assumption~\ref{assumption_5} and replacing $d\mathbf{X}$ by $d\|\mathbf{X}\|$ yields the desirable result.
\end{proof}
We next present the proof for Theorem~\ref{theorem_2}.
\begin{proof}
    Integrating Eq.~\ref{eq_22}:
    \begin{equation}
        \|\Delta\mathbf z(t)\|^2\leq \|\Delta\mathbf z(t_0)\|^2\text{exp}(-2\lambda\|\mathbf{X}\|_{TV,[0,t]}).
    \end{equation}
    Taking square roots and applying Assumption~\ref{assumption_2} to $\|\Delta\mathbf{z}(t_0)\|=\|\zeta_\phi(\hat{\mathbf{x}}_0)-\zeta_\phi(\hat{\mathbf{x}}'_0)\|$ yields the first part of the conclusion. Since $\|\mathbf{X}\|_{TV,[0,t]}\to\infty$ when $t\to\infty$, the limit is zero.
\end{proof}
The proof of Corollary~\ref{corollary_3} and Corollary~\ref{corollary_4} are presented as follows.
\begin{proof}
    Applying encoder and decoder Lipschitz continuities to Theorem~\ref{theorem_2} completes the proof.
\end{proof}
\begin{proof}
    Applying the second conclusion of Corollary~\ref{corollary_3} yields the desirable result.
\end{proof}
\section{Variable length \texorpdfstring{($\eta$)}{(eta)} and sparse \texorpdfstring{($\kappa$)}{(kappa)} control paths hyperparameters}

We discussed two regularization methods in section \ref{sec:reg_strat} and used those for synthetic, physics simulation, and forecasting datasets. This section contains the values of the hyperparameters for both regularization for all datasets. Table \ref{tab:synthetic_hyperparam} shows $\kappa$ \& $\eta$ for synthetic datasets. Only Sparse Control Paths regularization was used for all settings. Table \ref{tab:mujoco_hyperparam} shows $\kappa$ \& $\eta$ for physics simulation datasets. As discussed in the section \ref{sec:phy_data}, we used Variable Length Control Paths ($\eta$) regularization for the settings when the observed points are very few, like 20\%, or when Sparse Control Paths ($\kappa$) were not enough to provide generalization, e.g., \textsc{Hopper}.  

\begin{table}[H]
\centering
\small 
\caption{Hyperparameters $\kappa$ and $\eta$ for synthetic datasets in Table \ref{tab:fhn_glycolytic}}
\vspace{0.15cm}
\renewcommand{\arraystretch}{1.2}
\setlength{\tabcolsep}{6pt} 
\scalebox{0.85}{
\begin{tabular}{c|cc|cc|cc|cc}
\toprule
\% Observed Pts & \multicolumn{2}{c|}{20\%} & \multicolumn{2}{c|}{50\%} & \multicolumn{2}{c|}{80\%} & \multicolumn{2}{c}{100\%} \\
\midrule
Hyperparameters & $\kappa$ & $\eta$ & $\kappa$ & $\eta$ & $\kappa$ & $\eta$ & $\kappa$ & $\eta$ \\
\midrule
Lorenz            & 1.0 & 0 & 0.8 & 0 & 1.0 & 0 & 1.0 & 0 \\
Lotka Volterra    & 1.0 & 0 & 0.5 & 0 & 1.0 & 0 & 0.6 & 0 \\
FHN               & 1.0 & 0 & 0.4 & 0 & 0.7 & 0 & 0.6 & 0 \\
Glycolytic Oscillator & 1.0 & 0 & 1.0 & 0 & 0.5 & 0 & 0.5 & 0 \\
\bottomrule
\end{tabular}
}
\label{tab:synthetic_hyperparam}
\end{table}
\begin{table}[H]
\centering
\small 
\caption{Hyperparameters $\kappa$ and $\eta$ for MuJoCo datasets in Table \ref{tab:walker2d_pen}}
\vspace{0.15cm}
\renewcommand{\arraystretch}{1.2}
\setlength{\tabcolsep}{6pt} 
\scalebox{0.85}{
\begin{tabular}{c|cc|cc|cc|cc}
\toprule
\multirow{2}{*}{Dynamical System} & \multicolumn{2}{c|}{20\%} & \multicolumn{2}{c|}{50\%} & \multicolumn{2}{c|}{80\%} & \multicolumn{2}{c}{100\%} \\
\cline{2-9}
& $\kappa$ & $\eta$ & $\kappa$ & $\eta$ & $\kappa$ & $\eta$ & $\kappa$ & $\eta$ \\
\midrule
Hopper   & 1.0 & 10 & 1.0 & 10 & 0.2 & 10 & 0.2 & 10 \\
Walker2D & 1.0 & 10 & 0.6 & 0  & 0.6 & 0  & 0.7 & 0  \\
Pen      & 1.0 & 0  & 0.6 & 0  & 0.6 & 0  & 0.8 & 0  \\
Hammer   & 1.0 & 15 & 0.6 & 0  & 0.6 & 0  & 0.7 & 0  \\
\bottomrule
\end{tabular}
}
\label{tab:mujoco_hyperparam}
\end{table}

For long-term series forecasting (LTSF) problems, the hyperparameters are shown in Table \ref{tab:forecast_hyperparam_1} \& \ref{tab:forecast_hyperparam_2}. For most cases, we keep the values of $\kappa=0.7$ \& $\eta=10$ fixed except for Weather. 

\begin{table}[H]
\centering
\small 
\caption{Hyperparameters $\kappa$ and $\eta$ for LTSF datasets in Table \ref{tab:forecast_results}}
\vspace{0.15cm}
\renewcommand{\arraystretch}{1.2}
\setlength{\tabcolsep}{6pt} 
\scalebox{0.85}{
\begin{tabular}{c|cc|cc|cc|cc}
\toprule
Forecast Horizon ($T$) & \multicolumn{2}{c|}{96} & \multicolumn{2}{c|}{192} & \multicolumn{2}{c|}{336} & \multicolumn{2}{c}{720} \\
\midrule
Hyperparameters & $\kappa$ & $\eta$ & $\kappa$ & $\eta$ & $\kappa$ & $\eta$ & $\kappa$ & $\eta$ \\
\midrule
Exchange & 0.7 & 10 & 0.7 & 10 & 0.7 & 10 & 0.7 & 10 \\
ETTm2    & 0.7 & 10 & 0.7 & 10 & 0.7 & 10 & 0.7 & 10 \\
ETTh2    & 0.7 & 10 & 0.7 & 10 & 0.7 & 10 & 0.7 & 10 \\
Weather  & 1.0 & 0  & 0.7 & 50 & 1.0 & 50 & 1.0 & 0  \\
\bottomrule
\end{tabular}
}
\label{tab:forecast_hyperparam_1}
\end{table}

\begin{table}[H]
\centering
\small 
\caption{Hyperparameters $\kappa$ and $\eta$ for LTSF datasets in Table \ref{tab:forecast_results}}
\vspace{0.15cm}
\renewcommand{\arraystretch}{1.2}
\setlength{\tabcolsep}{6pt} 
\scalebox{0.85}{
\begin{tabular}{c|cc|cc|cc|cc}
\toprule
Forecast Horizon ($T$) & \multicolumn{2}{c|}{24} & \multicolumn{2}{c|}{36} & \multicolumn{2}{c|}{48} & \multicolumn{2}{c}{60} \\
\midrule
Hyperparameters & $\kappa$ & $\eta$ & $\kappa$ & $\eta$ & $\kappa$ & $\eta$ & $\kappa$ & $\eta$ \\
\midrule
ILI & 0.7 & 10 & 0.7 & 10 & 0.7 & 10 & 0.7 & 10 \\
\bottomrule
\end{tabular}
}
\label{tab:forecast_hyperparam_2}
\end{table}

\section{Time series forecasting}

Table~\ref{tab:forecast_results} in the main text reports LTSF results averaged over all forecast horizons, comparing four parameter-matched Correctors (MLP, GRU, Transformer, Neural CDE) against the frozen DLinear Predictor.
Table~\ref{tab:corrector_full} below provides the full per-horizon breakdown across all five datasets and all four forecast horizons.
The Neural CDE Corrector consistently improves DLinear across datasets and horizons except Weather $T$=720, where MLP achieves the lowest MSE/MAE.

\begin{table}[H]
\centering
\caption{Multivariate long-term forecasting errors (MSE/MAE; lower is better) for all forecast horizons across five LTSF datasets. All Correctors use the same frozen DLinear Predictor and are parameter-matched ($\approx$530\,K); architecture details are in Appendices~\ref{adx:discrete_baselines} and~\ref{adx:neural_cde}. Best value per row is highlighted in bold.}
\vspace{0.15cm}
\resizebox{0.95\textwidth}{!}{%
\begin{tabular}{c|c|cc|cc|cc|cc|cc}
\toprule
\multicolumn{2}{c|}{Methods}
& \multicolumn{2}{c|}{DLinear (w/o)}
& \multicolumn{2}{c|}{+MLP}
& \multicolumn{2}{c|}{+GRU}
& \multicolumn{2}{c|}{+Transformer}
& \multicolumn{2}{c}{+Neural CDE} \\
\midrule
\multicolumn{2}{c|}{Metric} & MSE & MAE & MSE & MAE & MSE & MAE & MSE & MAE & MSE & MAE \\
\midrule
\multirow{5}{*}{\rotatebox{90}{Exchange}}
& 96  & 0.086 & 0.211 & 0.345 & 0.462 & 0.325 & 0.451 & 0.188 & 0.335 & \textbf{0.083} & \textbf{0.207} \\
& 192 & 0.163 & 0.297 & 0.526 & 0.554 & 0.499 & 0.558 & 0.395 & 0.464 & \textbf{0.159} & \textbf{0.295} \\
& 336 & 0.334 & 0.443 & 0.975 & 0.780 & 0.721 & 0.673 & 0.564 & 0.571 & \textbf{0.243} & \textbf{0.370} \\
& 720 & 0.899 & 0.726 & 1.931 & 1.041 & 1.040 & 0.796 & 1.051 & 0.775 & \textbf{0.482} & \textbf{0.548} \\
& Avg & 0.370 & 0.419 & 0.944 & 0.709 & 0.646 & 0.619 & 0.549 & 0.536 & \textbf{0.242} & \textbf{0.355} \\
\midrule
\multirow{5}{*}{\rotatebox{90}{ETTm2}}
& 96  & 0.173 & 0.269 & 0.225 & 0.339 & 0.176 & 0.278 & 0.229 & 0.363 & \textbf{0.171} & \textbf{0.266} \\
& 192 & 0.239 & 0.315 & 0.365 & 0.441 & 0.276 & 0.353 & 0.273 & 0.360 & \textbf{0.235} & \textbf{0.311} \\
& 336 & 0.296 & 0.360 & 0.815 & 0.661 & 0.671 & 0.609 & 1.064 & 0.811 & \textbf{0.288} & \textbf{0.300} \\
& 720 & 0.427 & 0.440 & 2.956 & 1.247 & 2.817 & 1.200 & 2.859 & 1.313 & \textbf{0.389} & \textbf{0.422} \\
& Avg & 0.284 & 0.346 & 1.090 & 0.672 & 0.985 & 0.610 & 1.106 & 0.712 & \textbf{0.271} & \textbf{0.338} \\
\midrule
\multirow{5}{*}{\rotatebox{90}{ETTh2}}
& 96  & 0.293 & 0.355 & 1.200 & 0.745 & 0.505 & 0.543 & 0.895 & 0.748 & \textbf{0.290} & \textbf{0.353} \\
& 192 & 0.390 & 0.424 & 2.567 & 1.179 & 1.215 & 0.807 & 2.393 & 1.293 & \textbf{0.380} & \textbf{0.418} \\
& 336 & 0.464 & 0.474 & 4.230 & 1.701 & 2.813 & 1.205 & 2.521 & 1.352 & \textbf{0.458} & \textbf{0.470} \\
& 720 & 0.734 & 0.606 & 5.300 & 1.829 & 4.196 & 1.807 & 3.546 & 1.519 & \textbf{0.705} & \textbf{0.592} \\
& Avg & 0.470 & 0.465 & 3.324 & 1.363 & 2.182 & 1.091 & 2.339 & 1.228 & \textbf{0.458} & \textbf{0.458} \\
\midrule
\multirow{5}{*}{\rotatebox{90}{ILI}}
& 24  & 2.368 & 1.094 & 2.339 & 1.061 & 2.196 & 1.018 & \textbf{2.145} & \textbf{1.013} & 2.300 & 1.070 \\
& 36  & 2.291 & 1.080 & 2.474 & 1.072 & \textbf{2.190} & \textbf{1.032} & 2.237 & 1.047 & 2.250 & 1.040 \\
& 48  & 2.341 & 1.095 & 2.545 & 1.085 & \textbf{2.240} & \textbf{1.055} & 2.453 & 1.100 & 2.250 & 1.060 \\
& 60  & 2.611 & 1.170 & 2.833 & 1.164 & \textbf{2.566} & \textbf{1.142} & 2.965 & 1.274 & 2.590 & 1.150 \\
& Avg & 2.403 & 1.110 & 2.548 & 1.095 & \textbf{2.298} & \textbf{1.062} & 2.450 & 1.109 & 2.350 & 1.080 \\
\midrule
\multirow{5}{*}{\rotatebox{90}{Weather}}
& 96  & 0.176 & 0.235 & 0.181 & 0.261 & 0.172 & 0.242 & 0.175 & 0.268 & \textbf{0.159} & \textbf{0.219} \\
& 192 & 0.219 & 0.279 & 0.214 & 0.286 & 0.206 & 0.278 & 0.219 & 0.299 & \textbf{0.205} & \textbf{0.250} \\
& 336 & 0.263 & 0.315 & 0.271 & 0.325 & 0.264 & 0.325 & 0.276 & 0.339 & \textbf{0.251} & \textbf{0.300} \\
& 720 & 0.333 & 0.374 & \textbf{0.315} & \textbf{0.357} & 0.316 & 0.360 & 0.341 & 0.385 & 0.338 & 0.378 \\
& Avg & 0.248 & 0.301 & 0.245 & 0.307 & 0.240 & 0.301 & 0.253 & 0.323 & \textbf{0.238} & \textbf{0.286} \\
\bottomrule
\end{tabular}%
}
\label{tab:corrector_full}
\end{table}

\subsection{Full DLinear vs.\ transformer baselines}\label{adx:full_forecast_baselines}

The results for DLinear with Neural CDE Corrector against transformer-based forecasting baselines (TimesNet, FEDformer, Autoformer, Informer, Pyraformer, LogTrans) across all forecast horizons are listed in Table~\ref{tab:full_forecast}.

\begin{table}[H]
\caption{Multivariate long-term forecasting errors (MSE/MAE; lower is better). Best value in each row is highlighted in bold.}
\vspace{0.15cm}
\centering
\resizebox{\textwidth}{!}{
\begin{tabular}{c|c|cc|cc|cc|cc|cc|cc|cc|cc}
\toprule
\multicolumn{2}{c|}{Methods} 
& \multicolumn{2}{c|}{DLinear (w/)} 
& \multicolumn{2}{c|}{DLinear (w/o)} 
& \multicolumn{2}{c|}{TimesNet} 
& \multicolumn{2}{c|}{FEDformer} 
& \multicolumn{2}{c|}{Autoformer} 
& \multicolumn{2}{c|}{Informer} 
& \multicolumn{2}{c|}{Pyraformer} 
& \multicolumn{2}{c}{LogTrans} \\
\midrule
\multicolumn{2}{c|}{Metric} 
& MSE & MAE & MSE & MAE & MSE & MAE & MSE & MAE & MSE & MAE & MSE & MAE & MSE & MAE & MSE & MAE \\
\midrule
\multirow{5}{*}{\rotatebox{90}{Exchange}}
&96  & \textbf{0.083} & \textbf{0.207} & 0.085 & 0.210 & 0.107 & 0.234 & 0.148 & 0.278 & 0.197 & 0.323 & 0.847 & 0.752 & 0.376 & 1.105 & 0.968 & 0.812 \\
&192 & \textbf{0.159} & \textbf{0.295} & 0.162 & 0.297 & 0.226 & 0.344 & 0.271 & 0.380 & 0.300 & 0.369 & 1.204 & 0.895 & 1.748 & 1.151 & 1.040 & 0.851 \\
&336 & \textbf{0.243} & \textbf{0.370} & 0.333 & 0.442 & 0.367 & 0.448 & 0.460 & 0.500 & 0.509 & 0.524 & 1.672 & 1.036 & 1.874 & 1.172 & 1.659 & 1.081 \\
&720 & \textbf{0.482} & \textbf{0.548} & 0.896 & 0.724 & 0.964 & 0.746 & 1.195 & 0.841 & 1.447 & 0.941 & 2.478 & 1.310 & 1.943 & 1.206 & 1.941 & 1.127 \\
& Avg. & \textbf{0.242} & \textbf{0.355} & 0.369 & 0.418 & 0.416 & 0.443 & 0.518 & 0.500 & 0.613 & 0.539 & 1.550 & 0.998 & 1.485 & 1.159 & 1.402 & 0.968 \\
\midrule
\multirow{5}{*}{\rotatebox{90}{ETTm2}}
&96  & \textbf{0.171} & \textbf{0.266} & 0.173 & 0.268 & 0.187 & 0.267 & 0.203 & 0.287 & 0.255 & 0.339 & 0.365 & 0.453 & 0.435 & 0.507 & 0.768 & 0.642 \\
&192 & \textbf{0.235} & 0.311 & 0.239 & 0.315 & 0.249 & \textbf{0.309} & 0.269 & 0.328 & 0.281 & 0.340 & 0.533 & 0.563 & 0.730 & 0.673 & 0.989 & 0.757 \\
&336 & \textbf{0.288} & 0.354 & 0.295 & 0.359 & 0.321 & \textbf{0.351} & 0.325 & 0.366 & 0.339 & 0.372 & 1.363 & 0.887 & 1.201 & 0.845 & 1.334 & 0.872 \\
&720 & \textbf{0.389} & 0.422 & 0.426 & 0.439 & 0.408 & \textbf{0.403} & 0.421 & 0.415 & 0.433 & 0.432 & 3.379 & 1.338 & 3.625 & 1.451 & 3.048 & 1.328 \\
& Avg. & \textbf{0.271} & 0.338 & 0.283 & 0.345 & 0.291 & \textbf{0.333} & 0.304 & 0.349 & 0.327 & 0.371 & 1.410 & 0.810 & 1.498 & 0.869 & 1.535 & 0.900 \\
\midrule
\multirow{5}{*}{\rotatebox{90}{ETTh2}}
&96  & \textbf{0.290} & \textbf{0.353} & 0.292 & 0.354 & 0.340 & 0.374 & 0.346 & 0.388 & 0.358 & 0.397 & 3.755 & 1.525 & 0.645 & 0.597 & 2.116 & 1.197 \\
&192 & \textbf{0.380} & 0.418 & 0.388 & 0.422 & 0.402 & \textbf{0.414} & 0.429 & 0.439 & 0.456 & 0.452 & 5.602 & 1.931 & 0.788 & 0.683 & 4.315 & 1.635 \\
&336 & 0.458 & 0.470 & 0.466 & 0.475 & \textbf{0.452} & \textbf{0.452} & 0.496 & 0.487 & 0.482 & 0.486 & 4.721 & 1.835 & 0.907 & 0.747 & 1.124 & 1.604 \\
&720 & 0.705 & 0.592 & 0.729 & 0.604 & \textbf{0.462} & \textbf{0.468} & 0.463 & 0.474 & 0.515 & 0.511 & 3.647 & 1.625 & 0.963 & 0.783 & 3.188 & 1.540 \\
& Avg. & 0.458 & 0.458 & 0.469 & 0.464 & \textbf{0.414} & \textbf{0.427} & 0.433 & 0.447 & 0.453 & 0.462 & 4.431 & 1.729 & 0.826 & 0.703 & 2.686 & 1.494 \\
\midrule
\multirow{5}{*}{\rotatebox{90}{ILI}}
&24  & \textbf{2.30} & 1.07 & 2.36 & 1.09 & 2.317 & \textbf{0.934} & 3.228 & 1.260 & 3.483 & 1.287 & 5.764 & 1.677 & 1.420 & 2.012 & 4.480 & 1.444 \\
&36  & 2.25 & 1.04 & 2.28 & 1.07 & \textbf{1.972} & \textbf{0.920} & 2.679 & 1.080 & 3.103 & 1.148 & 4.755 & 1.467 & 7.394 & 2.031 & 4.799 & 1.467 \\
&48  & 2.25 & 1.06 & 2.35 & 1.09 & \textbf{2.238} & \textbf{0.940} & 2.622 & 1.078 & 2.669 & 1.085 & 4.763 & 1.469 & 7.551 & 2.057 & 4.800 & 1.468 \\
&60  & 2.59 & 1.15 & 2.61 & 1.17 & \textbf{2.027} & \textbf{0.928} & 2.857 & 1.157 & 2.770 & 1.125 & 5.264 & 1.564 & 7.662 & 2.100 & 5.278 & 1.560 \\
& Avg. & 2.35 & 1.08 & 2.40 & 1.10 & \textbf{2.139} & \textbf{0.931} & 2.846 & 1.144 & 3.006 & 1.161 & 5.136 & 1.544 & 6.007 & 2.050 & 4.839 & 1.485 \\
\midrule
\multirow{5}{*}{\rotatebox{90}{Weather}}
&96  & \textbf{0.159} & \textbf{0.219} & 0.175 & 0.235 & 0.172 & 0.220 & 0.217 & 0.296 & 0.266 & 0.336 & 0.300 & 0.384 & 0.896 & 0.556 & 0.458 & 0.490 \\
&192 & \textbf{0.205} & \textbf{0.250} & 0.218 & 0.278 & 0.219 & 0.261 & 0.276 & 0.336 & 0.307 & 0.367 & 0.598 & 0.544 & 0.622 & 0.624 & 0.658 & 0.589 \\
&336 & \textbf{0.251} & \textbf{0.300} & 0.263 & 0.314 & 0.280 & 0.306 & 0.339 & 0.380 & 0.359 & 0.395 & 0.578 & 0.523 & 0.739 & 0.753 & 0.797 & 0.652 \\
&720 & 0.338 & 0.378 & \textbf{0.332} & 0.374 & 0.365 & \textbf{0.359} & 0.403 & 0.428 & 0.419 & 0.428 & 1.059 & 0.741 & 1.004 & 0.934 & 0.869 & 0.675 \\
& Avg. & \textbf{0.238} & \textbf{0.286} & 0.247 & 0.300 & 0.259 & 0.287 & 0.309 & 0.360 & 0.338 & 0.382 & 0.634 & 0.548 & 0.815 & 0.717 & 0.696 & 0.601 \\
\bottomrule
\end{tabular}
}
\vspace{-0.2cm}

\label{tab:full_forecast}
\end{table}

\section{Implementation details}
This section discusses the training details of different Predictors, e.g., NODE, Contiformer, and DLinear, and the Corrector (Neural CDE).

\subsection{NODE}\label{adx:node_train}
The NODE was used as a Predictor for synthetic datasets. The vector field $\mathbf{f}$ (equation \ref{eq:ode_definition}) is approximated with a fully connected feedforward neural network. The hyperparameters are given below:

\begin{itemize}
    \item Batch size: 16
    \item Learning rate: 0.001
    \item Vector field $\mathbf{f}$: $\text{FC}(100)_{2}$
    \item Optimizer: $\mathtt{Adam}$
\end{itemize}
where $\text{FC}(100)_{2}$ denotes a fully connected feedforward neural network with 2 hidden layers each with 100 neurons.

\subsection{Contiformer}\label{adx:conti_train}
The Contiformer uses NODE to build a continuous-time transformer. A full connected feedforward neural network is used as an Encoder to learn the vector field. For integration, we used $\mathtt{odeint}$ from $\mathtt{torchdiffeq}$ \citep{torchdiffeq} with Dormand-Prince 5(4) (Dopri5), an adaptive explicit Runge-Kutta method, using relative/absolute tolerances ($\mathtt{rtol}=10^{-3}$, $\mathtt{atol}=10^{-6}$). The Contiformer has the following hyperparameters:

\begin{itemize}
    \item Batch size: 64
    \item Learning rate: 0.001
    \item Encoder: $\text{FC}(100)_{2}$
    \item Optimizer: $\mathtt{Adam}$
    \item Heads (H): 4
    \item Dimension of Key ($d_k$), Query ($d_q$), \& Value ($d_v$) per head  = 4
\end{itemize}

\subsection{DLinear}\label{adx:dlinear_train}
DLinear was proposed first in \citet{zeng2022transformerseffectivetimeseries} to challenge transformer-based solutions for LTSF datasets. We used the official implementation of DLinear from $\mathtt{Github}$ \footnote{\url{https://github.com/cure-lab/LTSF-Linear}} without changing any hyperparameters. This repository also includes LTSF datasets, which we used in our experiments. 

\subsection{Neural CDE} \label{adx:neural_cde}
The Neural CDE was used as a Corrector for all Predictors. We used the following hyperparameters for synthetic, physics simulation, and forecasting datasets.
\begin{itemize}
    \item Batch size: 256 except ILI where 32 was used 
    \item Learning rate: 0.001
    \item Neural CDE vector field $f_{\theta}$: $\text{FC}(400)_{4}$
    \item Neural CDE Decoder $\xi_{\varphi}$ (Appendix \ref{adx:decoder_size}): $\text{FC}(400)_{4}$
    \item Neural CDE initial hidden state network $\zeta_{\phi}$: $\text{FC}(50)_{1}$ 
    \item Neural CDE hidden state dimension $C$: 11 
    \item Optimizer: $\mathtt{Adam}$
\end{itemize}

The early stopping was used to stop the training of NODE, Contiformer, DLinear, and Neural CDE. The MSE loss function is used to train all Predictors. The solver settings for NODE and Neural CDE are the following. The $\mathtt{diffrax}$ implementation of $\mathtt{diffeqsolve}$ is used. For integration, we used $\mathtt{Tsit5}$ solver (Appendix \ref{adx:abl_ode_solvers}), a 5th-order explicit Runge-Kutta method with an embedded 4th-order method for adaptive step sizing. The PID controller, with relative/absolute tolerances ($\mathtt{rtol} = 10^{-3}$, $\mathtt{atol} = 10^{-6}$), is used to control the next step size based on the error estimate. The initial step size is 0.001. Experiments were run on machines with 32\,GB RAM and NVIDIA A100 GPUs.

\subsection{Continuous-time Corrector baselines}\label{adx:ct_baselines}
Latent ODE, SLCDE, and NRDE are natural continuous-time baselines for our Corrector because each models trajectories with differential-equation structure. They differ, however, in how directly the Predictor forecast path controls the residual dynamics. This distinction is important in the Predictor-Corrector setting: the target is not to model the data trajectory from scratch, but to model the residual trajectory induced by a fixed Predictor. We therefore want the Corrector hidden state to remain sensitive to local changes in the Predictor forecast path throughout the correction horizon.

Let $\mathbf{X}_t$ denote the interpolated Predictor forecast path constructed from $\hat{\mathbf{x}}_{0:T}$, and let $\mathbf{z}_t$ denote the Corrector hidden state. Neural CDE uses
\begin{equation}
    d\mathbf{z}_t = f_\theta(\mathbf{z}_t)\,d\mathbf{X}_t,\qquad \hat{\mathbf{e}}_t=\xi_\varphi(\mathbf{z}_t),
\end{equation}
so the hidden dynamics are continuously driven by local increments of the Predictor path. The baselines differ mainly in how they replace this controlled dynamics.

\textbf{Latent ODE} first encodes the observed forecast sequence into an initial latent state and then evolves NODE, e.g., $\dot{\mathbf{z}}_t=f_\theta(\mathbf{z}_t)$. This makes it suitable for latent trajectory modeling and irregularly sampled observations, but less aligned with residual correction because the Predictor path no longer drives the hidden dynamics after the initial encoding. If the Predictor error depends on local features of the forecast path, such as slope, phase drift, or curvature, an autonomous latent evolution must compress this information into the initial state rather than using it continuously.
Formally, it has the structure
\begin{equation}
    \mathbf{z}_{t_0}=\mathrm{Encoder}_\phi(\hat{\mathbf{x}}_{0:T}),\qquad
    \dot{\mathbf{z}}_t=f_\theta(\mathbf{z}_t),\qquad \hat{\mathbf{e}}_t=\xi_\varphi(\mathbf{z}_t),
\end{equation}
where $\mathbf{X}_t$ affects the dynamics only through the initial state.
\begin{itemize}
    \item Encoder: GRUCell with hidden state dimension $C = 11$
    \item Vector field $f_\theta$: $\text{FC}(400)_{4}$ (softplus activation)
    \item Decoder $\xi_\varphi$: $\text{FC}(400)_{4}$ (identical to Neural CDE)
    \item Batch size: 256
    \item Learning rate: 0.001
    \item Optimizer: $\mathtt{Adam}$
    \item Solver: $\mathtt{Tsit5}$
    \item Total parameters ($D=17$, $C=11$): $\approx$932K (dynamics $\approx$430K + encoder $\approx$10K + decoder $\approx$492K)
\end{itemize}

\textbf{SLCDE} is path-driven like Neural CDE, but uses a structured linear controlled dynamics model. It is therefore closer to our Corrector than Latent ODE because the hidden state is continuously driven by the control path. The main difference is expressivity: SLCDE imposes linear or structured dynamics, which can improve stability and efficiency, whereas our Neural CDE uses a nonlinear vector field to learn forecast-dependent residual dynamics. Thus, SLCDE is a useful constrained CDE-style baseline, while Neural CDE is the more flexible path-conditioned Corrector.
At a high level, SLCDE replaces the nonlinear vector field in Neural CDE with a structured linear controlled vector field,
\begin{equation}
    d\mathbf{z}_t = A_\theta(\mathbf{z}_t)\,d\mathbf{X}_t,\qquad \hat{\mathbf{e}}_t=\xi_\varphi(\mathbf{z}_t),
\end{equation}
where $A_\theta$ is constrained by the structured linear parameterization.
\begin{itemize}
    \item Vector field $A_\theta$: Linear($D+1$, $C \times C$), non-diagonal
    \item Decoder $\xi_\varphi$: $\text{FC}(400)_{4}$ (identical to Neural CDE)
    \item Hidden state dimension $C$: 11
    \item Batch size: 256
    \item Learning rate: 0.001
    \item Optimizer: $\mathtt{Adam}$
    \item Total parameters ($D=17$, $C=11$): $\approx$495K (vector field $(D+1){\times}C^2 = 2{,}178$ + decoder $\approx$492K); exact matching to Neural CDE ($\approx$1,057K) would require $C \approx 177$
\end{itemize}

\textbf{NRDE} augments the control path using rough-path features, such as log-signatures, to capture higher-order path interactions. This makes NRDE a strong path-driven baseline, especially when residuals depend on iterated or higher-order effects of the forecast path. In our experiments, NRDE was competitive only when log-signatures were computed on the finest available partition (implemented as stride 1); coarser partitions degraded performance. This fine-partition setting makes NRDE closest to Neural CDE's use of local path increments, but with a different path representation and computational profile.
Using a rough-path lift $\mathbf{X}_t$ of the Predictor path, NRDE can be viewed abstractly as
\begin{equation}
    d\mathbf{z}_t = g_\theta(\mathbf{z}_t)\,d\mathbf{X}_t,\qquad \hat{\mathbf{e}}_t=\xi_\varphi(\mathbf{z}_t),
\end{equation}
where $\mathbf{X}_t$ contains higher-order path information rather than only the original interpolated forecast path.
\begin{itemize}
    \item Log-signature depth: 2
    \item Partition: finest available (stride $= 1$)
    \item Vector field $g_\theta$: $\text{FC}(11)_{1}$
    \item Decoder $\xi_\varphi$: $\text{FC}(400)_{4}$ (identical to Neural CDE)
    \item Hidden state dimension $C$: 11
    \item Batch size: 256
    \item Learning rate: 0.001
    \item Optimizer: $\mathtt{Adam}$
    \item Solver: $\mathtt{rk4}$
    \item Total parameters ($D=17$, $C=11$): $\approx$515K (vector field $\approx$23K + decoder $\approx$492K); exact matching to Neural CDE ($\approx$1,057K) is impractical because the log-signature dimension $\ell_\text{sig}$ grows as $\mathcal{O}(D^2)$, making the vector field output $C \times \ell_\text{sig}$ grow rapidly with $D$ and producing degenerate architectures at higher dimensions
\end{itemize}

Overall, these baselines form a useful hierarchy. Latent ODE tests whether a continuous autonomous latent model is sufficient after encoding the forecast path. SLCDE tests whether a constrained path-driven CDE is sufficient. NRDE tests whether a rough-path representation of the forecast path improves correction. Neural CDE sits between these alternatives: it remains continuously conditioned on the Predictor path like SLCDE and NRDE, but uses a nonlinear learned vector field without requiring the same rough-path feature construction as NRDE.

\subsection{Discrete-time Corrector baselines}\label{adx:discrete_baselines}
For the LTSF experiments, DLinear is fixed as the Predictor and each Corrector receives the same forecast trajectory $\hat{\mathbf{x}}_{1:T}$, where $T$ is the forecast horizon and $\hat{\mathbf{x}}_t\in\mathbb{R}^{D}$ is the $D$-dimensional DLinear forecast at horizon step $t$. Let $\mathbf{x}_t\in\mathbb{R}^{D}$ denote the corresponding ground-truth value. The residual target is
\begin{equation}
    \mathbf{e}_t = \mathbf{x}_t-\hat{\mathbf{x}}_t,\qquad t=1,\ldots,T,
\end{equation}
and each Corrector $C_\theta$ maps the full forecast sequence to predicted residuals, $\hat{\mathbf{e}}_{1:T}=C_\theta(\hat{\mathbf{x}}_{1:T})$. The corrected forecast is then
\begin{equation}
    \hat{\mathbf{x}}^{\mathrm{corr}}_t=\hat{\mathbf{x}}_t+\hat{\mathbf{e}}_t.
\end{equation}
All discrete-time baselines are trained with the same residual loss and are parameter-matched to Neural CDE ($\approx$530\,K). Let $\mathbf{q}_t=[\hat{\mathbf{x}}_t,\tau_t]\in\mathbb{R}^{D+1}$ denote the forecast value concatenated with the normalized horizon index $\tau_t=t/T$.

\subsubsection{MLP Corrector}
The MLP Corrector applies the same pointwise network $g_\theta\colon\mathbb{R}^{D+1}\to\mathbb{R}^{D}$ independently at each horizon step:
\begin{equation}
    \hat{\mathbf{e}}_t=g_\theta(\mathbf{q}_t).
\end{equation}
This baseline tests whether local nonlinear correction of each forecast value is sufficient. It has no temporal memory and no interaction across horizon steps.
\begin{itemize}
    \item Hidden dimension: 512
    \item Layers: 3
    \item Optimizer: $\mathtt{Adam}$
\end{itemize}

\subsubsection{GRU Corrector}
The GRU Corrector introduces discrete sequential memory over the forecast horizon:
\begin{equation}
    \mathbf{h}_{1:T}=\mathrm{GRU}_\theta(\mathbf{q}_{1:T}),\qquad
    \hat{\mathbf{e}}_t=g_\phi(\mathbf{h}_t).
\end{equation}
Here $\mathbf{h}_t$ is the GRU hidden state after processing the forecast inputs up to step $t$, and $g_\phi$ is a decoder that maps this hidden state to a $D$-dimensional residual estimate. It is the closest discrete-time analogue to Neural CDE because both maintain a hidden state conditioned on the Predictor trajectory. The key difference is that GRU updates only at sampled forecast indices, while Neural CDE evolves continuously along the interpolated forecast path and is driven by its local increments.
\begin{itemize}
    \item Hidden dimension ($d_h$): 240
    \item Layers: 2
    \item Decoder: $\text{FC}(20)_{1}$
    \item Optimizer: $\mathtt{Adam}$
\end{itemize}

\subsubsection{Transformer Corrector}
The Transformer Corrector uses global self-attention over the full forecast sequence:
\begin{equation}
    \mathbf{H}_{1:T}=\mathrm{Transformer}_\theta(\mathbf{q}_{1:T}),\qquad
    \hat{\mathbf{e}}_t=W_{\mathrm{out}}\mathbf{H}_t+\mathbf{b}_{\mathrm{out}}.
\end{equation}
We implement this baseline as an encoder-only Transformer. This choice matches the residual-correction setting: the DLinear Predictor has already produced the full forecast horizon before the Corrector is applied, so the Corrector can attend to the entire sequence $\mathbf{q}_{1:T}$ at once rather than generating residuals autoregressively. The encoder produces one contextual representation $\mathbf{H}_t$ for each horizon step, and $W_{\mathrm{out}},\mathbf{b}_{\mathrm{out}}$ define the linear readout from this representation to the predicted residual.
\begin{itemize}
    \item Token embedding dimension ($d_\text{model}$): 128
    \item Attention heads ($h$): 4
    \item Feed-forward network (FFN) hidden dimension: 768
    \item Encoder layers: 2
    \item Optimizer: $\mathtt{Adam}$
\end{itemize}

\paragraph{Discrete versus continuous path conditioning.}
The three baselines isolate different discrete correction mechanisms: MLP tests pointwise residual calibration, GRU tests sequential discrete memory, and Transformer tests global discrete attention. Although GRU and Transformer are also conditioned on the Predictor forecasts, they use the forecasts as a discrete input sequence. Neural CDE instead treats the Predictor forecast as a continuous control path: its hidden state evolves between sampled forecast points and changes according to local path increments $d\mathbf{X}_t$. This gives the Corrector a continuous-time notion of how forecast geometry, such as slope and curvature, drives residual dynamics. This inductive bias may be especially useful for residual correction, which can differ from primary forecasting: the residual often reflects local error evolution, such as gradual drift, phase mismatch, or changes in trend, rather than a new long-horizon forecast to be generated from scratch. A Transformer can mix information globally across the forecast horizon, but global sequence mixing does not explicitly model how the residual state evolves along the path between neighboring forecast times. Under a matched parameter budget, Neural CDE may therefore allocate capacity more directly to the local evolution of forecast errors.

\section{Synthetic datasets}\label{adx:synth_data}
We train NODE on four synthetic datasets, shown in Table \ref{tab:fhn_glycolytic}. The closed-form expressions of multivariate ODEs are provided in this section. The parameters and initial conditions are adopted from \citet{shahid2025hopcast}.

\subsection{Lotka-Volterra \texorpdfstring{\citep{wangersky1978lotka}}{}}

\begin{align}
    \frac{dx}{dt} &= \alpha x - \beta xy \\
    \frac{dy}{dt} &= \delta xy - \gamma y
\end{align}

\textbf{Initial Condition Ranges}: $x \in [5,20]$; $y \in [5,10]$ \\
\textbf{Parameters}: $\alpha = 1.1$; $\beta = 0.4$; $\gamma = 0.4$; $\delta = 0.1$ 
 
\subsection{\textsc{Lorenz} \texorpdfstring{\citep{brunton2016discovering}}{}}

\begin{align}
    \frac{dx}{dt} &= \sigma (y-x) \\
    \frac{dy}{dt} &= x(\rho -z) - y \\
    \frac{dz}{dt} &= xy - \beta z
\end{align}
\textbf{Initial Condition Ranges}: $x \in [-20,20]$; $y \in [-20,20]$; $z \in [0,50]$ \\
\textbf{Parameters}: $\sigma = 10$; $\rho = 28$; $\beta = \frac{8}{3}$ 

\subsection{FitzHugh-Nagumo (\textsc{FHN}) \texorpdfstring{\citep{Izhikevich:2006}}{}}

\begin{align}
    \frac{dv}{dt} &= v - \frac{v^{3}}{3}-w+I\\
    \frac{dw}{dt} &= \epsilon(v+a-bw) 
\end{align}
\textbf{Initial Condition Ranges}: $v \in [-1.5,1.5]$;$w \in [-1.5,1.5]$ \\
\textbf{Parameters}: $a=0.7$;$b=0.8$;$\epsilon=0.08$;$I=0.5$ 

\subsection{\textsc{Glycolytic} Oscillator \texorpdfstring{\citep{daniels2015efficient}}{}}\label{adx:glycolytic}

\begin{align}
    \frac{dS_{1}}{dt} &= J_{0} - \frac{k_{1}S_{1}S_{6}}{1+(S_{6}/ K_{1})^{q}} \\
    \frac{dS_{2}}{dt} &= 2\frac{k_{1}S_{1}S_{6}}{1+(S_{6}/ K_{1})^{q}} -k_{2}S_{2}(N-S_{5}) - k_{6}S_{2}S_{5} \\
    \frac{dS_{3}}{dt} &= k_{2}S_{2}(N-S_{5})-k_{3}S_{3}(A-S_{6}) \\
    \frac{dS_{4}}{dt} &= k_{3}S_{3}(A-S_{6})-k_{4}S_{4}S_{5}-\kappa(S_{4}-S_{7})\\
    \frac{dS_{5}}{dt} &= k_{2}S_{2}(N-S_{5})-k_{4}S_{4}S_{5}-k_{6}S_{2}S_{5} \\
    \frac{dS_{6}}{dt} &= -2\frac{k_{1}S_{1}S_{6}}{1+(S_{6}/ K_{1})^{q}} +2k_{3}S_{3}(A-S_{6})-k_{5}S_{6} \\
    \frac{dS_{7}}{dt} &= \psi\kappa(S_{4}-S_{7})-kS_{7} 
\end{align}
\\
\textbf{Initial Condition Ranges}: $S_{1} \in [0.15,1.60];S_{2} \in [0.19,2.16];S_{3} \in [0.04,0.20];S_{4} \in [0.10,0.35]; S_{5} \in [0.08,0.30];S_{6} \in [0.14,2.67];S_{7} \in [0.05,0.10]$ \\
\textbf{Parameters}: $J_{0}=2.5; k_{1}=100; k_{2}=6; k_{3}=16; k_{4}=100; k_{5}=1.28; k_{6}=12$; $k=1.8; \kappa=13; q=4; K_{1}=0.52; \psi=0.1; N=1; A=4$

\subsection{Data generation details}

\begin{table}[H]
    \caption{Details about data generation}
    \vspace{0.15cm}
    \label{tab:data_detail}
    \centering
    \begin{tabular}{ccc|c|c|c|ccc}
    \toprule
    \multicolumn{3}{c|}{Model}
    & \multicolumn{1}{c}{$\Delta t$}
    & \multicolumn{1}{c}{Timesteps}
    & \multicolumn{1}{c}{Trajectories} \\
    \midrule
    \multicolumn{3}{c|}{\textsc{Lotka Volterra}}
    & \multicolumn{1}{c}{0.1}
    & \multicolumn{1}{c}{300}
    & \multicolumn{1}{c}{500} \\
    \midrule
    \multicolumn{3}{c|}{\textsc{Lorenz}}
    & \multicolumn{1}{c}{0.01}
    & \multicolumn{1}{c}{300}
    & \multicolumn{1}{c}{1000} \\
    \midrule
    \multicolumn{3}{c|}{\textsc{FHN}}
    & \multicolumn{1}{c}{0.5}
    & \multicolumn{1}{c}{400}
    & \multicolumn{1}{c}{350}\\
    \midrule
    \multicolumn{3}{c|}{\textsc{Glycolytic}}
    & \multicolumn{1}{c}{0.01}
    & \multicolumn{1}{c}{400}
    & \multicolumn{1}{c}{750} \\
    \bottomrule
    
    \end{tabular}
    
\end{table}

\section{Additional corrector results on \textsc{Lorenz} and \textsc{LVolt}}\label{adx:lorenz_lvolt}
Table~\ref{tab:lorenz_lvolt} extends the synthetic dataset evaluation from Table~1 of the main text to \textsc{Lorenz} and \textsc{LVolt}, reporting interpolation and extrapolation MSE for all four correctors across all sparsity levels.

\begin{table}[t]
\centering
\caption{%
  MSE without (w/o) and with (w/) corrector for \textsc{Lorenz} and \textsc{LVolt} (NODE predictor),
  combining interpolation ($0$--$50$) and extrapolation ($0$--$t^*$) regimes.
  $t^*$ is the maximum timestep sustaining $\geq 3\%$ cumulative MSE reduction.
  Best w/ per column in \textbf{bold} (interpolation); longest $t^*$ per column in
  \textbf{bold} (extrapolation).
  w/ values show mean${\scriptstyle\pm\text{std}}$ over 3 runs; w/o is deterministic (single Predictor).
}
\label{tab:lorenz_lvolt}
\setlength{\tabcolsep}{1pt}
\renewcommand{\arraystretch}{1.05}
\scriptsize
\resizebox{\textwidth}{!}{\begin{tabular}{l|l|l|c|c|c|c!{\vrule width 1.5pt}c|c|c|c}
\toprule
\multirow{2}{*}{\textbf{Dataset}} & \multirow{2}{*}{\textbf{Corrector}} & & \multicolumn{4}{c|}{\textbf{Interpolation} ($0$--$50$)} & \multicolumn{4}{c}{\textbf{Extrapolation} ($0$--$t^*$)} \\
\cline{4-11}
& & & 20\% & 50\% & 80\% & 100\% & 20\% & 50\% & 80\% & 100\% \\
\midrule
\multirow{12}{*}{\textsc{Lorenz}}
 & \multirow{3}{*}{SLCDE}
   & w/o & 2.308 & 1.061 & 1.062 & 0.8877 & 3.040 & 4.402 & 3.585 & 2.661 \\
 & & w/  & $1.989{\scriptstyle\pm.13}$ & $0.7812{\scriptstyle\pm.04}$ & $0.8049{\scriptstyle\pm.05}$ & $0.6495{\scriptstyle\pm.02}$ & $2.918{\scriptstyle\pm.24}$ & $4.256{\scriptstyle\pm.19}$ & $3.457{\scriptstyle\pm.21}$ & $2.561{\scriptstyle\pm.08}$ \\
 & & $0$--$t|\%$ & $0$--$50|14\%$ & $0$--$50|26\%$ & $0$--$50|24\%$ & $0$--$50|27\%$ & $0$--$75|4\%$ & $0$--$80|3\%$ & $0$--$75|4\%$ & $0$--$90|4\%$ \\
\cmidrule{2-11}
 & \multirow{3}{*}{Latent ODE}
   & w/o & 2.308 & 1.061 & 1.062 & 0.8877 & 3.040 & 4.002 & 2.923 & 1.159 \\
 & & w/  & $2.246{\scriptstyle\pm.17}$ & $0.9088{\scriptstyle\pm.06}$ & $0.8082{\scriptstyle\pm.04}$ & $0.7363{\scriptstyle\pm.03}$ & $2.930{\scriptstyle\pm.22}$ & $3.876{\scriptstyle\pm.25}$ & $2.777{\scriptstyle\pm.13}$ & $1.078{\scriptstyle\pm.06}$ \\
 & & $0$--$t|\%$ & $0$--$50|3\%$ & $0$--$50|14\%$ & $0$--$50|24\%$ & $0$--$50|17\%$ & $0$--$75|4\%$ & $0$--$75|3\%$ & $0$--$70|5\%$ & $0$--$70|7\%$ \\
\cmidrule{2-11}
 & \multirow{3}{*}{NRDE}
   & w/o & 2.308 & 1.061 & 1.062 & 0.8877 & 10.12 & 8.526 & 9.224 & 11.40 \\
 & & w/  & $\mathbf{0.4748}{\scriptstyle\pm.03}$ & $\mathbf{0.3236}{\scriptstyle\pm.01}$ & $0.3165{\scriptstyle\pm.02}$ & $\mathbf{0.2358}{\scriptstyle\pm.008}$ & $9.798{\scriptstyle\pm.81}$ & $8.235{\scriptstyle\pm.46}$ & $\mathbf{8.915}{\scriptstyle\pm.53}$ & $\mathbf{11.02}{\scriptstyle\pm.92}$ \\
 & & $0$--$t|\%$ & $0$--$50|79\%$ & $0$--$50|70\%$ & $0$--$50|70\%$ & $0$--$50|73\%$ & $0$--$175|3\%$ & $0$--$140|3\%$ & $\mathbf{0}$--$\mathbf{160}|3\%$ & $\mathbf{0}$--$\mathbf{180}|3\%$ \\
\cmidrule{2-11}
 & \multirow{3}{*}{Neural CDE}
   & w/o & 2.308 & 1.061 & 1.062 & 0.8877 & 12.13 & 10.25 & 8.652 & 7.878 \\
 & & w/  & $0.8948{\scriptstyle\pm.07}$ & $0.3717{\scriptstyle\pm.03}$ & $\mathbf{0.2890}{\scriptstyle\pm.01}$ & $0.4669{\scriptstyle\pm.04}$ & $11.50{\scriptstyle\pm1.02}$ & $9.939{\scriptstyle\pm.58}$ & $8.382{\scriptstyle\pm.31}$ & $7.638{\scriptstyle\pm.45}$ \\
 & & $0$--$t|\%$ & $0$--$50|61\%$ & $0$--$50|65\%$ & $0$--$50|73\%$ & $0$--$50|47\%$ & $\mathbf{0}$--$\mathbf{195}|5\%$ & $\mathbf{0}$--$\mathbf{165}|3\%$ & $0$--$155|3\%$ & $0$--$160|3\%$ \\
\midrule
\multirow{12}{*}{\textsc{LVolt}}
 & \multirow{3}{*}{SLCDE}
   & w/o & 0.1018 & 0.0357 & 0.0279 & 0.0253 & 0.2357 & 4.280 & 4.129 & 4.045 \\
 & & w/  & $0.0815{\scriptstyle\pm.006}$ & $0.0277{\scriptstyle\pm.002}$ & $0.0215{\scriptstyle\pm.001}$ & $0.0203{\scriptstyle\pm.001}$ & $0.2275{\scriptstyle\pm.018}$ & $4.119{\scriptstyle\pm.31}$ & $3.997{\scriptstyle\pm.27}$ & $3.905{\scriptstyle\pm.22}$ \\
 & & $0$--$t|\%$ & $0$--$50|20\%$ & $0$--$50|22\%$ & $0$--$50|23\%$ & $0$--$50|20\%$ & $0$--$80|3\%$ & $\mathbf{0}$--$\mathbf{105}|4\%$ & $\mathbf{0}$--$\mathbf{105}|3\%$ & $\mathbf{0}$--$\mathbf{105}|3\%$ \\
\cmidrule{2-11}
 & \multirow{3}{*}{Latent ODE}
   & w/o & 0.1018 & 0.0357 & 0.0279 & 0.0253 & --- & --- & --- & --- \\
 & & w/  & $0.1989{\scriptstyle\pm.019}$ & $0.1538{\scriptstyle\pm.008}$ & $0.1359{\scriptstyle\pm.011}$ & $0.1338{\scriptstyle\pm.005}$ & --- & --- & --- & --- \\
 & & $0$--$t|\%$ & $0$--$50|{-95}\%$ & $0$--$50|{-331}\%$ & $0$--$50|{-388}\%$ & $0$--$50|{-429}\%$ & $\leq 50$ & $\leq 50$ & $\leq 50$ & $\leq 50$ \\
\cmidrule{2-11}
 & \multirow{3}{*}{NRDE}
   & w/o & 0.1018 & 0.0357 & 0.0279 & 0.0253 & 0.7636 & 0.0500 & 0.0587 & 0.0931 \\
 & & w/  & $0.0629{\scriptstyle\pm.005}$ & $0.0274{\scriptstyle\pm.001}$ & $0.0220{\scriptstyle\pm.002}$ & $0.0206{\scriptstyle\pm.001}$ & $0.7401{\scriptstyle\pm.058}$ & $0.0464{\scriptstyle\pm.003}$ & $0.0551{\scriptstyle\pm.004}$ & $0.0894{\scriptstyle\pm.007}$ \\
 & & $0$--$t|\%$ & $0$--$50|38\%$ & $0$--$50|23\%$ & $0$--$50|21\%$ & $0$--$50|19\%$ & $\mathbf{0}$--$\mathbf{90}|3\%$ & $0$--$65|7\%$ & $0$--$70|6\%$ & $0$--$75|4\%$ \\
\cmidrule{2-11}
 & \multirow{3}{*}{Neural CDE}
   & w/o & 0.1018 & 0.0357 & 0.0279 & 0.0253 & 0.1531 & 0.2063 & 0.0318 & 0.0549 \\
 & & w/  & $\mathbf{0.0571}{\scriptstyle\pm.004}$ & $\mathbf{0.0241}{\scriptstyle\pm.001}$ & $\mathbf{0.0215}{\scriptstyle\pm.001}$ & $\mathbf{0.0193}{\scriptstyle\pm.001}$ & $0.1396{\scriptstyle\pm.012}$ & $0.1990{\scriptstyle\pm.016}$ & $0.0281{\scriptstyle\pm.002}$ & $0.0505{\scriptstyle\pm.004}$ \\
 & & $0$--$t|\%$ & $0$--$50|44\%$ & $0$--$50|32\%$ & $0$--$50|23\%$ & $0$--$50|24\%$ & $0$--$75|9\%$ & $0$--$80|4\%$ & $0$--$60|12\%$ & $0$--$70|8\%$ \\
\bottomrule
\end{tabular}}
\end{table}

\section{Physics simulation datasets}\label{adx:phy_sim}

\begin{figure}[ht]
    \centering
    \includegraphics[width=0.80\textwidth]{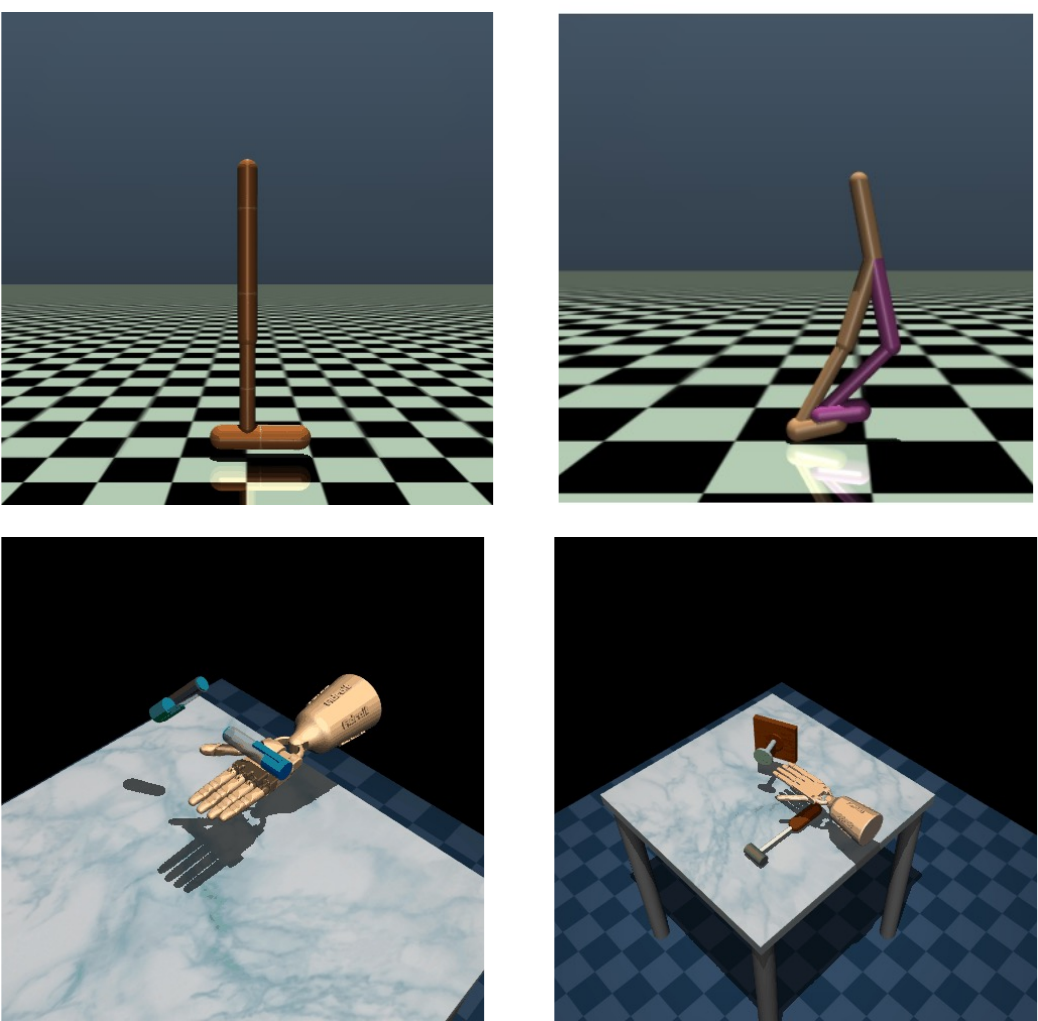}
    \caption{The MuJoCo environments inside the simulator \textbf{(a)} Upper left: \textsc{Hopper} (11D) \textbf{(b)} Upper right: \textsc{Walker2D} (17D) \textbf{(c)} Lower left: \textsc{Pen} (45D) \textbf{(d)} Lower right: \textsc{Hammer} (46D)}
    \label{fig:mujoco_phy}
\end{figure}

To collect the MuJoCo dataset, we trained expert policies for each environment and used them to generate trajectory rollouts. The policies were deterministic because the focus of this study is on learning the evolution of states in the system. Deterministic controllers ensure consistent state transitions across rollouts, whereas stochastic controllers could produce different trajectories from the same initial condition, which would complicate modeling when only the states are used as inputs. An extension to action-conditioned dynamics models is natural and would involve learning a mapping of the form $\mathbf{x}(t+1) = g_{\theta}(\mathbf{x}(t), \mathbf{a}(t))$ instead of $\mathbf{x}(t+1) = g_{\theta}(\mathbf{x}(t))$, where $\mathbf{x}(t+1)$ denotes the next state transitioned from $\mathbf{x}(t)$ by applying action $\mathbf{a}(t)$. Such a formulation would allow the use of data from arbitrary policies, including stochastic ones, and will be considered in future work. For each environment, we generated $2000$ trajectories of $300$ regularly-sampled time points each. There are four MuJoCo environments in Table.\ref{tab:walker2d_pen}. Fig. \ref{fig:mujoco_phy} shows the four environments within the simulator.

\section{Additional corrector results on \textsc{Hopper} and \textsc{Hammer}}\label{adx:hopper_hammer}
Table~\ref{tab:hopper_hammer} extends the physics simulation evaluation from Table~3 of the main text to \textsc{Hopper} and \textsc{Hammer}, reporting interpolation and extrapolation MSE for all four correctors across all sparsity levels.

\begin{table}[t]
\centering
\caption{%
  MSE without (w/o) and with (w/) corrector for \textsc{Hopper} and \textsc{Hammer}
  combining interpolation ($0$--$50$) and extrapolation ($0$--$t^*$) regimes.
  w/ values show mean${\scriptstyle\pm\text{std}}$ over 3 runs; w/o is deterministic (single Predictor).
}
\label{tab:hopper_hammer}
\setlength{\tabcolsep}{2pt}
\renewcommand{\arraystretch}{1.05}
\scriptsize
\resizebox{\textwidth}{!}{\begin{tabular}{l|l|l|c|c|c|c!{\vrule width 1.5pt}c|c|c|c}
\toprule
\multirow{2}{*}{\textbf{Dataset}} & \multirow{2}{*}{\textbf{Corrector}} & & \multicolumn{4}{c|}{\textbf{Interpolation} ($0$--$50$)} & \multicolumn{4}{c}{\textbf{Extrapolation} ($0$--$t^*$)} \\
\cline{4-11}
& & & 20\% & 50\% & 80\% & 100\% & 20\% & 50\% & 80\% & 100\% \\
\midrule
\multirow{12}{*}{\textsc{Hopper}}
 & \multirow{3}{*}{SLCDE}
   & w/o & 0.167 & 0.138 & 0.094 & 0.084 & 0.204 & 0.182 & 0.134 & 0.123 \\
 & & w/ & $0.063{\scriptstyle\pm.005}$ & $0.044{\scriptstyle\pm.003}$ & $0.031{\scriptstyle\pm.002}$ & $0.019{\scriptstyle\pm.001}$ & $0.140{\scriptstyle\pm.012}$ & $0.112{\scriptstyle\pm.008}$ & $0.109{\scriptstyle\pm.007}$ & $0.119{\scriptstyle\pm.009}$ \\
 & & $0$--$t^*|\%$ & $0$--$50|62\%$ & $0$--$50|68\%$ & $0$--$50|68\%$ & $0$--$50|78\%$ & $0$--$55|32\%$ & $0$--$55|38\%$ & $0$--$55|19\%$ & $0$--$55|3\%$ \\
\cmidrule{2-11}
 & \multirow{3}{*}{Latent ODE}
   & w/o & 0.167 & 0.138 & 0.094 & 0.084 & 0.204 & 0.182 & 0.134 & 0.123 \\
 & & w/ & $\mathbf{0.013}{\scriptstyle\pm.001}$ & $0.012{\scriptstyle\pm.0008}$ & $0.013{\scriptstyle\pm.001}$ & $0.008{\scriptstyle\pm.0004}$ & $0.118{\scriptstyle\pm.011}$ & $0.088{\scriptstyle\pm.005}$ & $0.097{\scriptstyle\pm.008}$ & $0.109{\scriptstyle\pm.006}$ \\
 & & $0$--$t^*|\%$ & $0$--$50|92\%$ & $0$--$50|92\%$ & $0$--$50|86\%$ & $0$--$50|91\%$ & $0$--$55|43\%$ & $0$--$55|52\%$ & $0$--$55|28\%$ & $0$--$55|12\%$ \\
\cmidrule{2-11}
 & \multirow{3}{*}{NRDE}
   & w/o & 0.167 & 0.138 & 0.094 & 0.084 & 0.204 & 0.182 & 0.134 & 0.123 \\
 & & w/ & $0.014{\scriptstyle\pm.001}$ & $\mathbf{0.006}{\scriptstyle\pm.0004}$ & $\mathbf{0.006}{\scriptstyle\pm.0003}$ & $\mathbf{0.005}{\scriptstyle\pm.0002}$ & $0.107{\scriptstyle\pm.009}$ & $0.112{\scriptstyle\pm.010}$ & $0.084{\scriptstyle\pm.006}$ & $0.094{\scriptstyle\pm.007}$ \\
 & & $0$--$t^*|\%$ & $0$--$50|92\%$ & $0$--$50|96\%$ & $0$--$50|94\%$ & $0$--$50|95\%$ & $0$--$55|48\%$ & $0$--$55|38\%$ & $0$--$55|38\%$ & $0$--$55|23\%$ \\
\cmidrule{2-11}
 & \multirow{3}{*}{Neural CDE}
   & w/o & 0.167 & 0.138 & 0.094 & 0.084 & 0.385 & 0.172 & 0.142 & 0.136 \\
 & & w/ & $0.061{\scriptstyle\pm.005}$ & $0.018{\scriptstyle\pm.002}$ & $0.035{\scriptstyle\pm.003}$ & $0.028{\scriptstyle\pm.001}$ & $0.372{\scriptstyle\pm.031}$ & $0.149{\scriptstyle\pm.009}$ & $0.120{\scriptstyle\pm.013}$ & $0.132{\scriptstyle\pm.007}$ \\
 & & $0$--$t^*|\%$ & $0$--$50|63\%$ & $0$--$50|86\%$ & $0$--$50|63\%$ & $0$--$50|67\%$ & $\mathbf{0}$--$\mathbf{110}|4\%$ & $\mathbf{0}$--$\mathbf{65}|13\%$ & $\mathbf{0}$--$\mathbf{70}|15\%$ & $\mathbf{0}$--$\mathbf{60}|3\%$ \\
\midrule
\multirow{12}{*}{\textsc{Hammer}}
 & \multirow{3}{*}{SLCDE}
   & w/o & 0.020 & 0.014 & 0.011 & 0.009 & 0.0204 & 0.0131 & 0.0096 & 0.0080 \\
 & & w/ & $0.009{\scriptstyle\pm.0007}$ & $0.009{\scriptstyle\pm.0005}$ & $0.007{\scriptstyle\pm.0003}$ & $0.007{\scriptstyle\pm.0002}$ & $0.0181{\scriptstyle\pm.0016}$ & $0.0125{\scriptstyle\pm.0007}$ & $0.0092{\scriptstyle\pm.0006}$ & $0.0077{\scriptstyle\pm.0003}$ \\
 & & $0$--$t^*|\%$ & $0$--$50|57\%$ & $0$--$50|38\%$ & $0$--$50|32\%$ & $0$--$50|22\%$ & $0$--$65|11\%$ & $0$--$75|5\%$ & $0$--$85|3\%$ & $0$--$75|4\%$ \\
\cmidrule{2-11}
 & \multirow{3}{*}{Latent ODE}
   & w/o & 0.020 & 0.014 & 0.011 & 0.009 & 0.0208 & 0.0133 & 0.0104 & 0.0082 \\
 & & w/ & $0.007{\scriptstyle\pm.0006}$ & $0.007{\scriptstyle\pm.0004}$ & $0.008{\scriptstyle\pm.0005}$ & $0.006{\scriptstyle\pm.0002}$ & $0.0137{\scriptstyle\pm.0011}$ & $0.0121{\scriptstyle\pm.0009}$ & $0.0096{\scriptstyle\pm.0006}$ & $0.0075{\scriptstyle\pm.0004}$ \\
 & & $0$--$t^*|\%$ & $0$--$50|63\%$ & $0$--$50|47\%$ & $0$--$50|30\%$ & $0$--$50|31\%$ & $0$--$60|34\%$ & $0$--$60|10\%$ & $0$--$60|8\%$ & $0$--$60|9\%$ \\
\cmidrule{2-11}
 & \multirow{3}{*}{NRDE}
   & w/o & 0.020 & 0.014 & 0.011 & 0.009 & 0.0205 & 0.0076 & 0.0065 & 0.0053 \\
 & & w/ & $\mathbf{0.003}{\scriptstyle\pm.0003}$ & $\mathbf{0.003}{\scriptstyle\pm.0001}$ & $\mathbf{0.003}{\scriptstyle\pm.0001}$ & $\mathbf{0.003}{\scriptstyle\pm.0001}$ & $0.0167{\scriptstyle\pm.0014}$ & $0.0073{\scriptstyle\pm.0004}$ & $0.0062{\scriptstyle\pm.0005}$ & $0.0050{\scriptstyle\pm.0003}$ \\
 & & $0$--$t^*|\%$ & $0$--$50|84\%$ & $0$--$50|78\%$ & $0$--$50|73\%$ & $0$--$50|70\%$ & $0$--$70|19\%$ & $0$--$145|4\%$ & $0$--$135|4\%$ & $0$--$130|5\%$ \\
\cmidrule{2-11}
 & \multirow{3}{*}{Neural CDE}
   & w/o & 0.020 & 0.014 & 0.011 & 0.009 & 0.0158 & 0.0073 & 0.0049 & 0.0046 \\
 & & w/ & $0.012{\scriptstyle\pm.0009}$ & $0.006{\scriptstyle\pm.0004}$ & $0.004{\scriptstyle\pm.0002}$ & $0.003{\scriptstyle\pm.0001}$ & $0.0152{\scriptstyle\pm.0013}$ & $0.0070{\scriptstyle\pm.0004}$ & $0.0047{\scriptstyle\pm.0004}$ & $0.0043{\scriptstyle\pm.0003}$ \\
 & & $0$--$t^*|\%$ & $0$--$50|37\%$ & $0$--$50|56\%$ & $0$--$50|64\%$ & $0$--$50|63\%$ & $\mathbf{0}$--$\mathbf{100}|4\%$ & $\mathbf{0}$--$\mathbf{150}|4\%$ & $\mathbf{0}$--$\mathbf{180}|4\%$ & $\mathbf{0}$--$\mathbf{150}|6\%$ \\
\bottomrule
\end{tabular}}
\end{table}

\section{Ablation studies}

\subsection{Training horizon for LTSF}\label{adx:forecast_corr_hor}
In section \ref{sec:fore_data}, we discussed that the Corrector is trained on the first 50, 100, 150, and 300 timesteps for forecast horizons 96, 192, 336, and 720, respectively. Here, we chose forecast horizon 336 for Exchange, ETTm2, ETTh2, \& Weather to demonstrate the impact of training Corrector on the first 50, 100, 150, 200, 250, 300, and 336 timesteps. Table \ref{tab:train_hor_forecast} shows that the Corrector shows a relatively poor performance, i.e., an increase in MSE/MAE, when trained on very short (e.g., 50 timesteps) or very long horizons (e.g., 336 timesteps). This motivates our choice of training the Corrector on intermediate-length horizons (e.g., 50 for $T$=96, 100 for $T$=192, etc.) to report results in Table \ref{tab:forecast_results} and \ref{tab:full_forecast}.

\begin{table}[H]
\centering
\small 
\caption{Multivariate LTSF errors (MSE/MAE) for forecast horizon 336 of Exchange, ETTm2, ETTh2, \& Weather, where the Corrector is trained on the first 50, 100, 150, 200, 250, 300, \& 336 timesteps for each setting. The \%$\downarrow$ shows the reduction in MSE in percentage.}
\vspace{0.15cm}
\renewcommand{\arraystretch}{1.2}
\setlength{\tabcolsep}{4pt} 
\begin{adjustbox}{max width=\textwidth} 
\begin{tabular}{c|cc|cc|cc|cc|cc|cc|cc}
\toprule
Train Horizon & \multicolumn{2}{c|}{50} & \multicolumn{2}{c|}{100} & \multicolumn{2}{c|}{150} & \multicolumn{2}{c|}{200} & \multicolumn{2}{c|}{250} & \multicolumn{2}{c|}{300} & \multicolumn{2}{c}{336} \\ 
\cline{0-14}
Metrics & MSE & MAE & MSE & MAE & MSE & MAE & MSE & MAE & MSE & MAE & MSE & MAE & MSE & MAE \\
\midrule
\multirow{2}{*}{Exchange} & 0.304 & 0.413  & 0.278 & 0.394 & 0.243 & 0.370 & 0.305 & 0.434 & 0.353 & 0.470 & 0.340 & 0.460 & 0.360 & 0.472 \\
& 8.71\%$\downarrow$ & 6.56\%$\downarrow$ & 16.51\%$\downarrow$ & 10.86\%$\downarrow$ & 27.03\%$\downarrow$ & 16.29\%$\downarrow$ & 8.41\%$\downarrow$ & 1.81\%$\downarrow$ & -6.01\%$\downarrow$ & -6.3\%$\downarrow$ & -2.10\%$\downarrow$ & -4.07\%$\downarrow$ & -8.11\%$\downarrow$ & -6.79\%$\downarrow$ \\
\midrule
\multirow{2}{*}{ETTh2} & 0.463 & 0.473 & 0.459 & 0.470 & 0.458 & 0.470 & 0.456 & 0.469 & 0.459 & 0.470 & 0.460 & 0.471 & 0.471 & 0.480 \\
  & 0.64\%$\downarrow$ & 0.42\%$\downarrow$ & 1.50\%$\downarrow$ & 1.05\%$\downarrow$ & 1.72\%$\downarrow$ & 1.05\%$\downarrow$ & 2.15\%$\downarrow$ & 1.26\%$\downarrow$  & 1.50\%$\downarrow$ & 1.05\%$\downarrow$ & 1.29\%$\downarrow$ & 0.84\%$\downarrow$ & -1.07\%$\downarrow$ & -1.05\%$\downarrow$ \\
\midrule
\multirow{2}{*}{ETTm2} & 0.288 & 0.353 & 0.287 & 0.352 & 0.288 & 0.354 & 0.290 & 0.356 & 0.293 & 0.358 & 0.291 & 0.355 & 0.292 & 0.357 \\
 & 2.37\%$\downarrow$ & 1.67\%$\downarrow$ & 2.71\%$\downarrow$ & 1.95\%$\downarrow$ & 2.37\%$\downarrow$ & 1.39\%$\downarrow$ & 1.69\%$\downarrow$ & 0.84\%$\downarrow$ & 0.68\%$\downarrow$ & 0.28\%$\downarrow$ & 0.68\%$\downarrow$ & 0.28\%$\downarrow$ & 1.02\%$\downarrow$ & 0.56\%$\downarrow$ \\
\midrule
\multirow{2}{*}{Weather}  & 0.264 & 0.320 & 0.253 & 0.306 & 0.251 & 0.300 & 0.253 & 0.305 & 0.254 & 0.303 & 0.256 & 0.306 & 0.251 & 0.300 \\
& -0.38\%$\downarrow$ & -1.91\%$\downarrow$ & 3.80\%$\downarrow$ & 2.55\%$\downarrow$ & 4.56\%$\downarrow$ & 3.82\%$\downarrow$ & 3.80\%$\downarrow$ & 2.87\%$\downarrow$ & 3.42\%$\downarrow$ & 4.08\%$\downarrow$ & 3.42\%$\downarrow$ & 2.55\%$\downarrow$ & 4.56\%$\downarrow$ & 4.46\%$\downarrow$ \\
\bottomrule
\end{tabular}
\end{adjustbox}
\label{tab:train_hor_forecast}
\end{table}

\subsection{Order of ODE}

\rev{To show the robustness of Predictor-Corrector to different orders of ODEs. We generated synthetic data from second, third, and fourth-order ODEs.}

\rev{\paragraph{Second-order system.}
We simulate the damped oscillator
\begin{equation}
    x''(t) + 0.3\,x'(t) + 1.0\,x(t) = 0.
\end{equation}
Its first-order state-space representation is
\begin{equation}
\begin{aligned}
    z_1 &= x, \\
    z_2 &= x', \\
    \dot z_1 &= z_2, \\
    \dot z_2 &= -1.0\,z_1 - 0.3\,z_2.
\end{aligned}
\end{equation}}

\rev{\paragraph{Third-order system.}
We simulate the third-order linear ODE
\begin{equation}
    x^{(3)}(t) + 0.4\,x''(t) + 0.3\,x'(t) + 1.0\,x(t) = 0.
\end{equation}
Its first-order state-space form is
\begin{equation}
\begin{aligned}
    z_1 &= x, \\
    z_2 &= x', \\
    z_3 &= x'', \\
    \dot z_1 &= z_2, \\
    \dot z_2 &= z_3, \\
    \dot z_3 &= -1.0\,z_1 - 0.3\,z_2 - 0.4\,z_3.
\end{aligned}
\end{equation}}

\rev{\paragraph{Fourth-order system.}
We simulate the fourth-order ODE
\begin{equation}
    x^{(4)}(t) + 0.3\,x^{(3)}(t) + 0.5\,x''(t) + 0.3\,x'(t) + 1.0\,x(t) = 0.
\end{equation}
The corresponding first-order state-space form is
\begin{equation}
\begin{aligned}
    z_1 &= x, \\
    z_2 &= x', \\
    z_3 &= x'', \\
    z_4 &= x^{(3)}, \\
    \dot z_1 &= z_2, \\
    \dot z_2 &= z_3, \\
    \dot z_3 &= z_4, \\
    \dot z_4 &= -1.0\,z_1 - 0.3\,z_2 - 0.5\,z_3 - 0.3\,z_4.
\end{aligned}
\end{equation}}

\rev{We train NODE as a Predictor for these examples with the same architectural details given in \ref{adx:node_train}. Both states and derivatives are modeled with NODE. The Corrector is trained with the same architectural details given in \ref{adx:neural_cde}. To simulate time-varying partial observability, we randomly mask half of the observed features for 0\%, 50\%, and 80\% of the observed time points. The results are shown in the Table. \ref{tab:order_ode}. The interpolation shows the performance of the Corrector for the first 50 timesteps, which corresponds to its training horizon. The extrapolation columns lists the timesteps for every setting up to which the Corrector brings at least a 3\% reduction in MSE of NODE. The corrector consistently brings a reduction in the MSE of the Predictor irrespective of the levels of partial observability. However, the order of the ODE does impact the performance of the Corrector. The Corrector brings less than 10\% reduction in MSE within the interpolation region for the 4th order system compared to the 2nd and 3rd order systems, where the reduction in MSEs is significantly higher. Within the extrapolation region, the horizons up to which the Corrector brings at least 3\% reduction in MSE are much smaller for 4th order systems compared to 2nd and 3rd order systems. This demonstrates that the order of the system impacts the performance of the Corrector.}

\begin{table*}[t]
\centering
\small 
\caption{Test MSE of NODE as a Predictor on ODEs of different orders with (w/) and without (w/o) Corrector. The $\% \downarrow$ shows the percentage reduction in MSE of NODE with our proposed Corrector. For both interpolation and extrapolation, reported MSE values are computed from timestep 0 up to the specified timestep ($t$) for each setting (0-$t$).}
\vspace{0.05cm}
\renewcommand{\arraystretch}{1.2}
\setlength{\tabcolsep}{4pt} 
\begin{adjustbox}{max width=0.9\textwidth} 
\begin{tabular}{c|c|ccc|ccc}
\toprule
\multicolumn{1}{c|}{\multirow{2}{*}{\parbox[c]{2cm}{\centering \textbf{Dynamical}\\ \textbf{System}}}} & 
\multicolumn{1}{c|}{\multirow{2}{*}{\textbf{Model}}} &
\multicolumn{3}{c|}{\textbf{Interpolation} (\% Pts w/ missing features)} &
\multicolumn{3}{c}{\textbf{Extrapolation} (\% Pts w/ missing features)} \\
\cline{3-8} 
& & 0\% & 30\% & 60\% & 0\% & 30\% & 60\% \\
\midrule
\multicolumn{1}{c}{\multirow{2}{*}{\textbf{2nd Order ODE}}}
& \multicolumn{1}{|c|}{w/o}
& $0.0142$ & $0.0452$ & $0.0728$ & $0.0413$ & $0.0952$ & $0.109$ \\
\cline{2-8}
& \multicolumn{1}{|c|}{w/}
& $0.0039$ & $0.0152$ & $0.0252$ & $0.0397$ & $0.0922$ & $0.097$ \\
\cline{2-8}
& \multicolumn{1}{|c|}{$0$ -- $t$ $\boldsymbol{|}$ \%$\downarrow$} 
& $0$--$50$ $\boldsymbol{|}$ $72\%$ & $0$--$50$ $\boldsymbol{|}$ $66\%$ & $0$--$50$ $\boldsymbol{|}$ $65\%$ & $0$--$170$ $\boldsymbol{|}$ $4\%$ & $0$--$180$ $\boldsymbol{|}$ $3\%$ & $0$--$165$ $\boldsymbol{|}$ $10\%$ \\
\midrule 
\multicolumn{1}{c}{\multirow{2}{*}{\textbf{3rd Order ODE}}}
& \multicolumn{1}{|c|}{w/o}
& $0.0101$ & $0.0359$ & $0.0934$ & $7.3$ & $8.1$ & $9.4$ \\
\cline{2-8}
& \multicolumn{1}{|c|}{w/}
& $0.0019$ & $0.0039$ & $0.0543$ & $6.9$ & $7.9$ & $9.0$ \\
\cline{2-8}
& \multicolumn{1}{|c|}{$0$ -- $t$ $\boldsymbol{|}$ \%$\downarrow$} 
& $0$--$50$ $\boldsymbol{|}$ $81\%$ & $0$--$50$ $\boldsymbol{|}$ $89\%$ & $0$--$50$ $\boldsymbol{|}$ $41\%$ & $0$--$180$ $\boldsymbol{|}$ $5\%$ & $0$--$170$ $\boldsymbol{|}$ $4\%$ & $0$--$190$ $\boldsymbol{|}$ $4\%$ \\
\midrule
\multicolumn{1}{c}{\multirow{2}{*}{\textbf{4th Order ODE}}}
& \multicolumn{1}{|c|}{w/o}
& $1.27$ & $2.3$ & $4.1$ & $10.36$ & $12.45$ & $14.56$ \\
\cline{2-8}
& \multicolumn{1}{|c|}{w/}
& $1.15$ & $2.1$ & $3.8$ & $10.02$ & $11.9$ & $14.10$ \\
\cline{2-8}
& \multicolumn{1}{|c|}{$0$ -- $t$ $\boldsymbol{|}$ \%$\downarrow$} 
& $0$--$50$ $\boldsymbol{|}$ $9\%$ & $0$--$50$ $\boldsymbol{|}$ $8\%$ & $0$--$50$ $\boldsymbol{|}$ $7\%$ & $0$--$90$ $\boldsymbol{|}$ $3\%$ & $0$--$80$ $\boldsymbol{|}$ $4\%$ & $0$--$100$ $\boldsymbol{|}$ $3\%$ \\
\bottomrule
\end{tabular}
\end{adjustbox}
\label{tab:order_ode}
\end{table*}

\subsection{Sensitivity to interpolation schemes}\label{adx:abl_inter_schemes}

\rev{The $\mathtt{diffrax}$ \cite{kidger2022neuraldifferentialequations} contains different interpolation schemes for control paths of Neural CDE. Here, we present the sensitivity of Corrector performance to those interpolation schemes. The results with Cubic interpolation for \textsc{Walker2D} and \textsc{Hammer} are copied from the Table \ref{tab:walker2d_pen}. The Linear interpolation brings a slightly smaller reduction in MSE (\%) compared to the Cubic interpolation in almost every setting in Table \ref{tab:interp_scheme}. This observation is in line with the results reported by \citet{morrill2022choice}.}

\begin{table*}[t]
\centering
\small 
\caption{MSE of ContiFormer as a Predictor on various dynamical systems’ datasets from MuJoCo \citep{todorov2012mujoco} with (w/) and without (w/o) Corrector under different interpolation schemes for control paths of Neural CDE. For interpolation and extrapolation, the reported MSE values are computed from timestep $0$ up to the specified timestep ($t$) for each setting (0-$t$).}
\vspace{0.05cm}
\renewcommand{\arraystretch}{1.2}
\setlength{\tabcolsep}{4pt} 
\begin{adjustbox}{max width=0.9\textwidth} 
\begin{tabular}{c|c|cccc|cccc}
\toprule
\multicolumn{1}{c|}{\multirow{2}{*}{\parbox[c]{2.5cm}{\centering \textbf{Dynamical System}\\ (Interpolation)}}} & 
\multicolumn{1}{c|}{\multirow{2}{*}{\textbf{Model}}} &
\multicolumn{4}{c|}{\textbf{Interpolation} (\% Observed Pts)} &
\multicolumn{4}{c}{\textbf{Extrapolation} (\% Observed Pts)} \\
\cline{3-10} 
& & 20\% & 50\% & 80\% & 100\% & 20\% & 50\% & 80\% & 100\% \\
\midrule
\multirow{2}{*}{\textbf{Walker2D} (Linear)}
& \multicolumn{1}{|c|}{w/o}
& 1.826 & 0.572 & 0.408 & 0.164 & 2.72 & 0.577 & 0.778 & 0.869 \\
\cline{2-10}
& \multicolumn{1}{|c|}{w/}
& 1.029 & 0.435 & 0.228 & 0.073 & 2.62 & 0.532 & 0.758 & 0.839 \\
\cline{2-10}
& \multicolumn{1}{|c|}{$0$--$t$ $\boldsymbol{|}$ \%$\downarrow$} 
& $0$--$50$ $\boldsymbol{|}$ $43\%$ & $0$--$50$ $\boldsymbol{|}$ $23\%$ & $0$--$50$ $\boldsymbol{|}$ $43\%$ & $0$--$50$ $\boldsymbol{|}$ $55\%$ & $0$--$180$ $\boldsymbol{|}$ $4\%$ & $0$--$100$ $\boldsymbol{|}$ $7\%$ & $0$--$130$ $\boldsymbol{|}$ $3\%$ & $0$--$140$ $\boldsymbol{|}$ $3\%$ \\
\midrule 
\multirow{2}{*}{\textbf{Walker2D} (Cubic)}
& \multicolumn{1}{|c|}{w/o}
& 1.826 & 0.572 & 0.408 & 0.164 & 2.89 & 1.70 & 0.778 & 0.869 \\
\cline{2-10}
& \multicolumn{1}{|c|}{w/}
& 0.721 & 0.285 & 0.149 & 0.053 & 2.68 & 1.63 & 0.750 & 0.838 \\
\cline{2-10}
& \multicolumn{1}{|c|}{$0$--$t$ $\boldsymbol{|}$ \%$\downarrow$} 
& $0$--$50$ $\boldsymbol{|}$ $61\%$ & $0$--$50$ $\boldsymbol{|}$ $50\%$ & $0$--$50$ $\boldsymbol{|}$ $63\%$ & $0$--$50$ $\boldsymbol{|}$ $68\%$ 
& $0$--$190$ $\boldsymbol{|}$ $7\%$ & $0$--$180$ $\boldsymbol{|}$ $4\%$ & $0$--$130$ $\boldsymbol{|}$ $4\%$ & $0$--$140$ $\boldsymbol{|}$ $4\%$ \\
\midrule
\multirow{2}{*}{\textbf{Hammer} (Linear)}
& \multicolumn{1}{|c|}{w/o}
& 0.020 & 0.0137 & 0.0106 & 0.0085 & 0.0115 & 0.0083 & 0.0049 & 0.0046 \\
\cline{2-10}
& \multicolumn{1}{|c|}{w/}
& 0.0122 & 0.0058 & 0.0038 & 0.0039 & 0.0110 & 0.0080 & 0.0047 & 0.0045 \\
\cline{2-10}
& \multicolumn{1}{|c|}{$0$--$t$ $\boldsymbol{|}$ \%$\downarrow$} 
& $0$--$50$ $\boldsymbol{|}$ $39\%$ & $0$--$50$ $\boldsymbol{|}$ $57\%$ & $0$--$50$ $\boldsymbol{|}$ $63\%$ & $0$--$50$ $\boldsymbol{|}$ $54\%$ 
& $0$--$140$ $\boldsymbol{|}$ $4\%$ & $0$--$140$ $\boldsymbol{|}$ $3\%$ & $0$--$180$ $\boldsymbol{|}$ $4\%$ & $0$--$150$ $\boldsymbol{|}$ $6\%$ \\
\midrule
\multirow{2}{*}{\textbf{Hammer} (Cubic)}
& \multicolumn{1}{|c|}{w/o}
& 0.020  & 0.0137 & 0.0106 & 0.0085 & 0.0158 & 0.0073  & 0.0049  & 0.0046 \\
\cline{2-10}
& \multicolumn{1}{|c|}{w/}
& 0.012 & 0.0059 & 0.0037 & 0.0032 & 0.0150 & 0.0070 & 0.0047 & 0.0044 \\
\cline{2-10}
& \multicolumn{1}{|c|}{$0$--$t$ $\boldsymbol{|}$ \%$\downarrow$} 
& $0$--$50$ $\boldsymbol{|}$ $37\%$ & $0$--$50$ $\boldsymbol{|}$ $56\%$ & $0$--$50$ $\boldsymbol{|}$ $64\%$ & $0$--$50$ $\boldsymbol{|}$ $63\%$ 
& $0$--$100$ $\boldsymbol{|}$ $4\%$ & $0$--$150$ $\boldsymbol{|}$ $4\%$ & $0$--$180$ $\boldsymbol{|}$ $4\%$ & $0$--$150$ $\boldsymbol{|}$ $6\%$ \\
\bottomrule
\end{tabular}
\end{adjustbox}
\label{tab:interp_scheme}
\end{table*}

\subsection{Sensitivity to ODE solvers}\label{adx:abl_ode_solvers}

\rev{The $\mathtt{diffrax}$ package has different explicit Runga-Kutta (RK) methods. We used $\mathtt{Tsit5}$ to report results in the paper everywhere else. Here, we analyze the sensitivity of Walker2D results to other solvers from the explicit RK family. The results are reported in Table \ref{tab:ode_solvers}. The $\mathtt{Tsit5}$ performed better both in interpolation and extrapolation regions compared to other solvers, followed by $\mathtt{Dopri5}$. The adaptive-step size solvers, such as $\mathtt{Heun}$, $\mathtt{Dopri5}$, \& $\mathtt{Tsit5}$, performed better than $\mathtt{Euler}$ method.}

\begin{table}[H]
\centering
\small 
\caption{MSE of ContiFormer as a Predictor on Walker2D dataset from MuJoCo \citep{todorov2012mujoco} with (w/) and without (w/o) Corrector under different solvers for Neural CDE. For interpolation and extrapolation, the reported MSE values are computed from timestep $0$ up to the specified timestep ($t$) for each setting (0-$t$).}
\vspace{0.05cm}
\renewcommand{\arraystretch}{1.2}
\setlength{\tabcolsep}{4pt} 
\begin{adjustbox}{max width=\textwidth} 
\begin{tabular}{c|c|cccc|cccc}
\toprule
\multicolumn{1}{c|}{\multirow{2}{*}{\parbox[c]{2.5cm}{\centering \textbf{ODE Solver}}}} & 
\multicolumn{1}{c|}{\multirow{2}{*}{\textbf{Model}}} &
\multicolumn{4}{c|}{\textbf{Interpolation} (\% Observed Pts)} &
\multicolumn{4}{c}{\textbf{Extrapolation} (\% Observed Pts)} \\
\cline{3-10} 
& & 20\% & 50\% & 80\% & 100\% & 20\% & 50\% & 80\% & 100\% \\
\midrule
\multirow{2}{*}{$\mathtt{Euler}$}
& \multicolumn{1}{|c|}{w/o}
& 1.826 & 0.572 & 0.408 & 0.164 & 2.72 & 0.639 & 0.778 & 0.245 \\
\cline{2-10}
& \multicolumn{1}{|c|}{w/}
& 1.375 & 0.437 & 0.292 & 0.105 & 2.62 & 0.620 & 0.743 & 0.234 \\
\cline{2-10}
& \multicolumn{1}{|c|}{$0$--$t$ $\boldsymbol{|}$ \%$\downarrow$} 
& $0$--$50$ $\boldsymbol{|}$ $24\%$ & $0$--$50$ $\boldsymbol{|}$ $23\%$ & $0$--$50$ $\boldsymbol{|}$ $28\%$ & $0$--$50$ $\boldsymbol{|}$ $36\%$ & $0$--$180$ $\boldsymbol{|}$ $4\%$ & $0$--$100$ $\boldsymbol{|}$ $3\%$ & $0$--$130$ $\boldsymbol{|}$ $4\%$ & $0$--$80$ $\boldsymbol{|}$ $4\%$ \\
\midrule 
\multirow{2}{*}{$\mathtt{Heun}$}
& \multicolumn{1}{|c|}{w/o}
& 1.826 & 0.572 & 0.408 & 0.164 & 1.86 & 0.969 & 0.778 & 0.413 \\
\cline{2-10}
& \multicolumn{1}{|c|}{w/}
& 0.901 & 0.359 & 0.328 & 0.078 & 1.67 & 0.925 & 0.752 & 0.394 \\
\cline{2-10}
& \multicolumn{1}{|c|}{$0$--$t$ $\boldsymbol{|}$ \%$\downarrow$} 
& $0$--$50$ $\boldsymbol{|}$ $50\%$ & $0$--$50$ $\boldsymbol{|}$ $37\%$ & $0$--$50$ $\boldsymbol{|}$ $19\%$ & $0$--$50$ $\boldsymbol{|}$ $52\%$ 
& $0$--$130$ $\boldsymbol{|}$ $9\%$ & $0$--$130$ $\boldsymbol{|}$ $4\%$ & $0$--$130$ $\boldsymbol{|}$ $4\%$ & $0$--$120$ $\boldsymbol{|}$ $4\%$ \\
\midrule
\multirow{2}{*}{$\mathtt{Dopri5}$}
& \multicolumn{1}{|c|}{w/o}
& 1.826 & 0.572 & 0.408 & 0.164 & 1.86 & 0.969 & 0.778 & 0.869 \\
\cline{2-10}
& \multicolumn{1}{|c|}{w/}
& 0.819 & 0.361 & 0.329 & 0.064 & 1.81 & 0.929 & 0.760 & 0.841 \\
\cline{2-10}
& \multicolumn{1}{|c|}{$0$--$t$ $\boldsymbol{|}$ \%$\downarrow$} 
& $0$--$50$ $\boldsymbol{|}$ $55\%$ & $0$--$50$ $\boldsymbol{|}$ $36\%$ & $0$--$50$ $\boldsymbol{|}$ $19\%$ & $0$--$50$ $\boldsymbol{|}$ $60\%$ 
& $0$--$130$ $\boldsymbol{|}$ $3\%$ & $0$--$130$ $\boldsymbol{|}$ $4\%$ & $0$--$130$ $\boldsymbol{|}$ $3\%$ & $0$--$140$ $\boldsymbol{|}$ $4\%$ \\
\midrule
\multirow{2}{*}{$\mathtt{Tsit5}$}
& \multicolumn{1}{|c|}{w/o}
& 1.826 & 0.572 & 0.408 & 0.164 & 2.89 & 1.70 & 0.778 & 0.869 \\
\cline{2-10}
& \multicolumn{1}{|c|}{w/}
& 0.721 & 0.285 & 0.149 & 0.053 & 2.68 & 1.63 & 0.750 & 0.838 \\
\cline{2-10}
& \multicolumn{1}{|c|}{$0$--$t$ $\boldsymbol{|}$ \%$\downarrow$} 
& $0$--$50$ $\boldsymbol{|}$ $61\%$ & $0$--$50$ $\boldsymbol{|}$ $50\%$ & $0$--$50$ $\boldsymbol{|}$ $63\%$ & $0$--$50$ $\boldsymbol{|}$ $68\%$ 
& $0$--$190$ $\boldsymbol{|}$ $7\%$ & $0$--$180$ $\boldsymbol{|}$ $4\%$ & $0$--$130$ $\boldsymbol{|}$ $4\%$ & $0$--$140$ $\boldsymbol{|}$ $4\%$ \\
\bottomrule
\end{tabular}
\end{adjustbox}
\label{tab:ode_solvers}
\end{table}

\subsection{Decoder \texorpdfstring{($\xi_{\varphi}$)}{(xi)} size}\label{adx:decoder_size}

\rev{The decoder ($\xi_{\varphi}$) for Neural CDE corrector maps the hidden state dynamics $\mathbf{z}(t)$ to error dynamics $\mathbf{\hat{e}}(t)$ as shown in Fig. \ref{fig:method_fig}. Here, we show the results of \textsc{FHN} with varying decoder sizes in the Table. \ref{tab:decoder_size}. The results elsewhere are reported using a four-layer fully connected neural network with 400 neurons in each layer, denoted as $\text{FC}(400)_{4}$. With $\text{FC}(20)_{1}$, we observe a performance degradation within the interpolation region, but the extrapolation performance is better. This indicates that a small decoder mitigates overfitting and facilitates generalization, while the hidden state $\mathbf{z}(t)$ models error dynamics. }

\begin{table}[H]
\centering
\small 
\caption{MSE of NODE as a Predictor on FHN dataset with (w/) and without (w/o) Corrector under different decoder sizes for Neural CDE (Corrector). For interpolation and extrapolation, the reported MSE values are computed from timestep $0$ up to the specified timestep ($t$) for each setting (0-$t$).}
\vspace{0.05cm}
\renewcommand{\arraystretch}{1.2}
\setlength{\tabcolsep}{4pt} 
\begin{adjustbox}{max width=\textwidth} 
\begin{tabular}{c|c|cccc|cccc}
\toprule
\multicolumn{1}{c|}{\multirow{2}{*}{\parbox[c]{2.5cm}{\centering \textbf{Decoder Size}}}} & 
\multicolumn{1}{c|}{\multirow{2}{*}{\textbf{Model}}} &
\multicolumn{4}{c|}{\textbf{Interpolation} (\% Observed Pts)} &
\multicolumn{4}{c}{\textbf{Extrapolation} (\% Observed Pts)} \\
\cline{3-10} 
& & 20\% & 50\% & 80\% & 100\% & 20\% & 50\% & 80\% & 100\% \\
\midrule
\multicolumn{1}{c}{\multirow{2}{*}{$\text{FC}(20)_{1}$}}
& \multicolumn{1}{|c|}{w/o}
& $0.225$ & $0.161$ & $0.150$ & $0.137$ & $0.264$ & $0.183$ & $0.200$ & $0.180$ \\
\cline{2-10}
& \multicolumn{1}{|c|}{w/}
& $0.187$ & $0.139$ & $0.122$ & $0.114$ & $0.248$ & $0.177$ & $0.191$ & $0.171$ \\
\cline{2-10}
& \multicolumn{1}{|c|}{$0$ -- $t$ $\boldsymbol{|}$ \%$\downarrow$} 
& $0$--$50$ $\boldsymbol{|}$ $17\%$ & $0$--$50$ $\boldsymbol{|}$ $14\%$ & $0$--$50$ $\boldsymbol{|}$ $18\%$ & $0$--$50$ $\boldsymbol{|}$ $17\%$ & $0$--$90$ $\boldsymbol{|}$ $5\%$ & $0$--$190$ $\boldsymbol{|}$ $4\%$ & $0$--$220$ $\boldsymbol{|}$ $5\%$ & $0$--$200$ $\boldsymbol{|}$ $5\%$ \\
\midrule
\multicolumn{1}{c}{\multirow{2}{*}{$\text{FC}(100)_{1}$}}
& \multicolumn{1}{|c|}{w/o}
& $0.225$ & $0.161$ & $0.150$ & $0.137$ & $0.274$ & $0.178$ & $0.192$ & $0.179$ \\
\cline{2-10}
& \multicolumn{1}{|c|}{w/}
& $0.181$ & $0.137$ & $0.123$ & $0.107$ & $0.263$ & $0.169$ & $0.168$ & $0.173$ \\
\cline{2-10}
& \multicolumn{1}{|c|}{$0$ -- $t$ $\boldsymbol{|}$ \%$\downarrow$} 
& $0$--$50$ $\boldsymbol{|}$ $18\%$ & $0$--$50$ $\boldsymbol{|}$ $12\%$ & $0$--$50$ $\boldsymbol{|}$ $18\%$ & $0$--$50$ $\boldsymbol{|}$ $22\%$ & $0$--$100$ $\boldsymbol{|}$ $4\%$ & $0$--$150$ $\boldsymbol{|}$ $4\%$ & $0$--$200$ $\boldsymbol{|}$ $4\%$ & $0$--$180$ $\boldsymbol{|}$ $4\%$ \\
\midrule 
\multicolumn{1}{c}{\multirow{2}{*}{$\text{FC}(400)_{4}$}}
& \multicolumn{1}{|c|}{w/o}
& $0.225$ & $0.161$ & $0.150$ & $0.137$ & $0.242$ & $0.178$ & $0.192$ & $0.166$ \\
\cline{2-10}
& \multicolumn{1}{|c|}{w/}
& $0.164$ & $0.128$ & $0.093$ & $0.097$ & $0.231$ & $0.171$ & $0.183$ & $0.158$ \\
\cline{2-10}
& \multicolumn{1}{|c|}{$0$ -- $t$ $\boldsymbol{|}$ \%$\downarrow$} 
& $0$--$50$ $\boldsymbol{|}$ $27\%$ & $0$--$50$ $\boldsymbol{|}$ $20\%$ & $0$--$50$ $\boldsymbol{|}$ $38\%$ & $0$--$50$ $\boldsymbol{|}$ $29\%$ & $0$--$75$ $\boldsymbol{|}$ $5\%$ & $0$--$150$ $\boldsymbol{|}$ $4\%$ & $0$--$140$ $\boldsymbol{|}$ $4\%$ & $0$--$150$ $\boldsymbol{|}$ $5\%$ \\
\bottomrule
\end{tabular}
\end{adjustbox}
\label{tab:decoder_size}
\end{table}

\subsection{Efficiency--extrapolation Pareto curves for \texorpdfstring{$\kappa$}{kappa} and \texorpdfstring{$\eta$}{eta} regularization}\label{adx:pareto_curves}

\begin{figure}[H]
    \centering
    \includegraphics[width=1.0\textwidth]{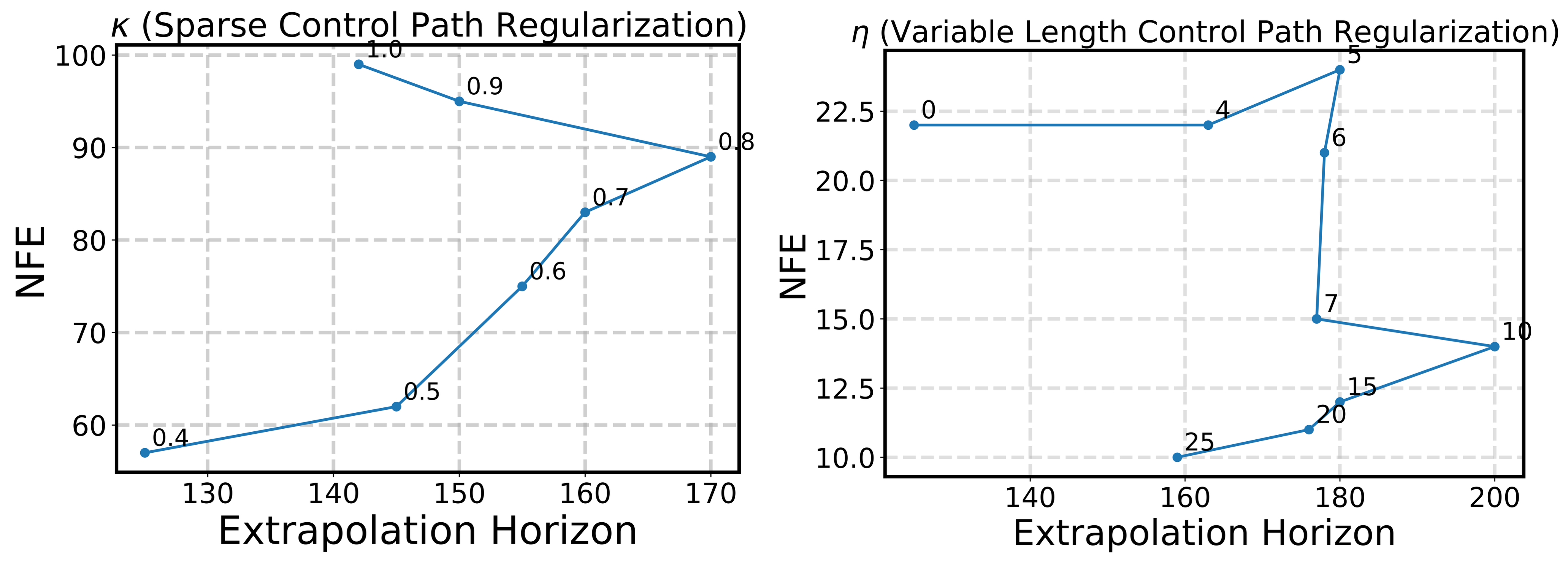}
    \caption{Pareto curves showing the trade-offs between extrapolation and efficiency via the proposed regularization strategies.}
    \label{fig:pareto_curves}
\end{figure}

\rev{The $\kappa$ (Sparse Control Paths) and $\eta$ (Variable Length Control Paths) regularization strategies are two proposed approaches to enhance the extrapolation performance and computational efficiency (i.e., NFE) of the Neural CDE corrector. Both approaches offer a trade-off between the efficiency, i.e.,  number of function evaluations (NFEs), and the extrapolation horizon. The extrapolation horizon is the timestep up to which the corrector brings at least 3\% reduction in MSE. We plot Pareto curves for both $\kappa$ and $\eta$ to show the trade-off between NFE and extrapolation horizon in Fig. \ref{fig:pareto_curves}. Pareto curves for $\kappa$ \& $\eta$ show the value of $\kappa$ \& $\eta$ right next to each data point, while the x and y axes show the NFE and extrapolation horizon, respectively. The curves show that reducing NFE and extending the extrapolation horizon are competing objectives, so the preferred setting is the point that best balances both quantities.}
 
\subsection{Wall-clock time of \texorpdfstring{$\kappa$}{kappa} \& \texorpdfstring{$\eta$}{eta}}\label{adx:clock_time}

\begin{figure}[H]
    \centering
    \includegraphics[width=1.0\textwidth]{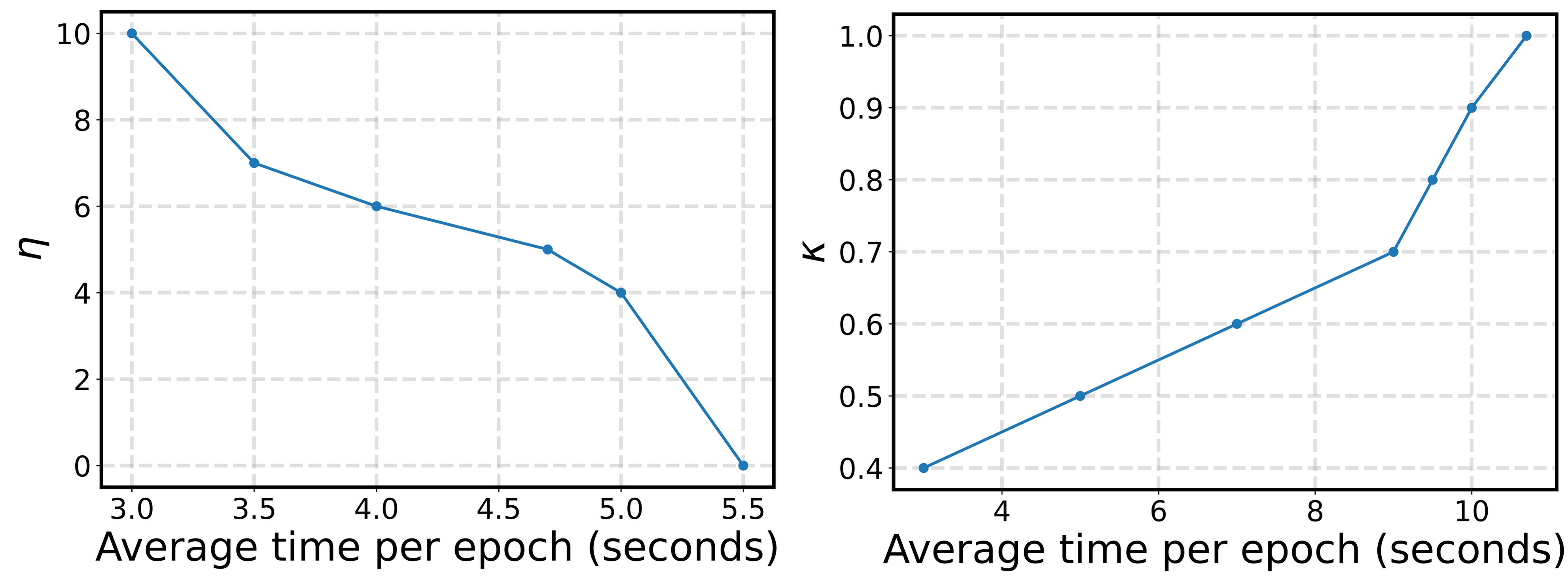}
    \caption{Average wall-clock time (in seconds) of an epoch with varying values of $\kappa$ \& $\eta$.}
    \label{fig:kappa_eta_clock}
\end{figure}

\rev{We report the average wall-clock time it takes to complete one epoch during training, varying the levels of $\kappa$ \& $\eta$, as shown in Fig. \ref{fig:kappa_eta_clock}. It can be seen that smaller values of $\kappa$ and larger values of $\eta$ result in faster training, corroborating the results shown in Fig. \ref{fig:nfe_reduction} of the paper.}

\subsection{Long-horizon stress tests}\label{adx:stress_test}

\begin{figure}[H]
    \centering
    \includegraphics[width=0.55\textwidth]{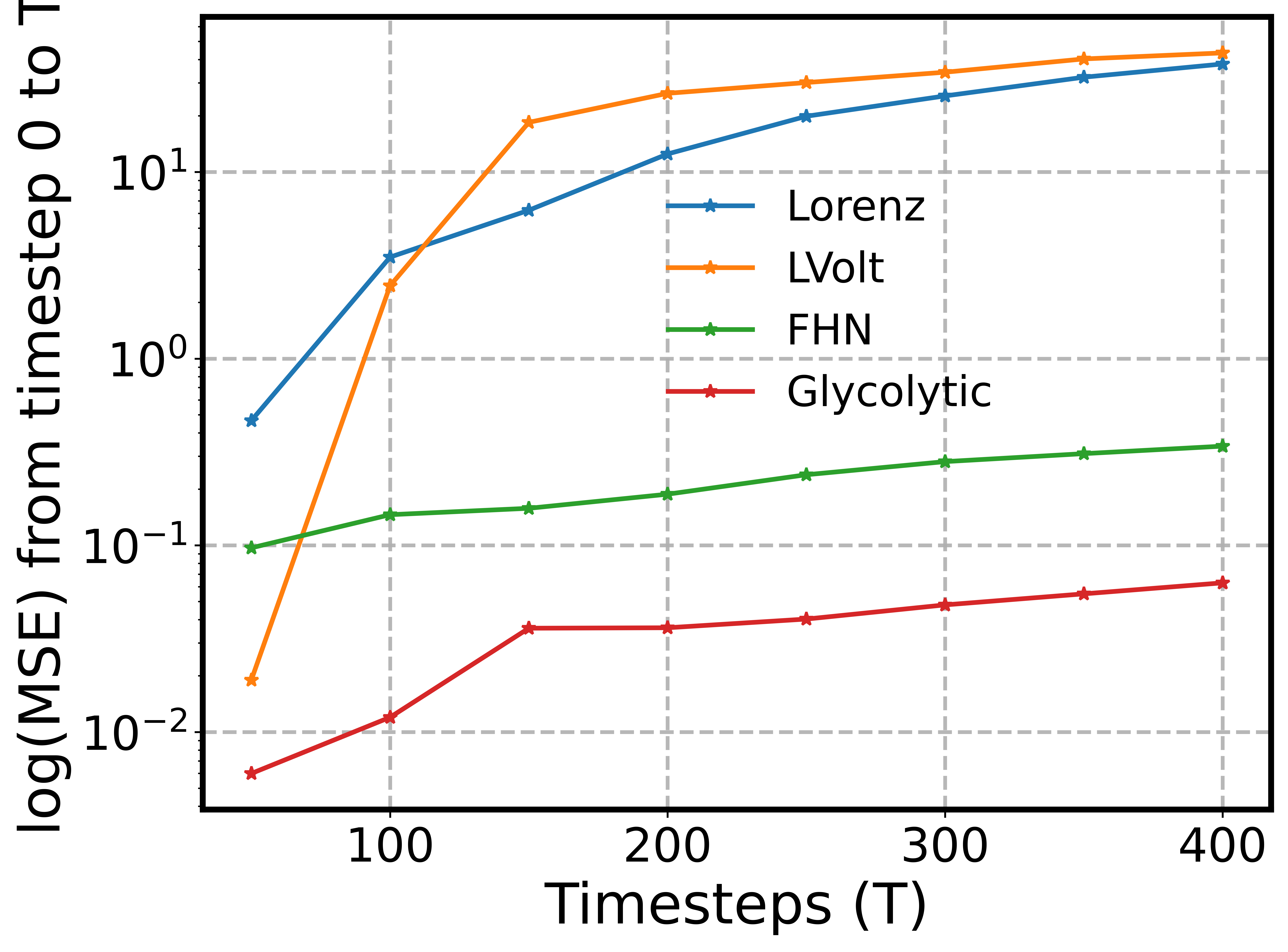}
    \caption{The long-horizon tests demonstrating the well-boundedness of the error of the corrected forecasts of the Predictor (NODE) on \textsc{Lorenz}, \textsc{LVolt}, \textsc{FHN}, \& \textsc{Glycolytic}.}
    \label{fig:long_horizon_stress_test}
\end{figure}

\rev{We empirically show that the MSE of corrected forecasts of Predictor over long horizons (400 timesteps) remains well-bounded. The results are shown for \textsc{Lorenz}, \textsc{FHN}, \textsc{LVolt}, and \textsc{Glycolytic} in the Fig. \ref{fig:long_horizon_stress_test}. Each data point shows the log(MSE) from timestep 0 to timestep T on the x-axis.}

\section{Additional LTSF results}

%
%

\subsection{Performance of Neural CDE Corrector on FEDformer}\label{adx:ncde_fedformer}

\rev{To demonstrate the effectiveness of Neural CDE corrector on a transformer-based model for LTSF, we test Neural CDE to improve the performance of FEDformer on the ETTm2 dataset. The results are shown in the Table. \ref{tab:fedformer_ltsf}.}

\begin{table}[H]
\caption{Multivariate long-term forecasting errors (MSE/MAE; lower is better). The results of FEDformer with (w/) and without (w/o) our Corrector on ETTm2 are shown.}
\vspace{0.15cm}
\centering
\begin{adjustbox}{width=0.4\textwidth, center}
\begin{tabular}{c|c|cc|cc}
\toprule
\multicolumn{2}{c|}{Methods} 
& \multicolumn{2}{c|}{\textbf{FEDformer (w/)}}
& \multicolumn{2}{c}{\textbf{FEDformer (w/o)}}  \\
\midrule
\multicolumn{2}{c|}{Metric} 
& MSE & MAE & MSE & MAE \\
\midrule
\multirow{5}{*}{\rotatebox{90}{ETTm2}}
&96  & 0.189 & 0.265 & 0.203 & 0.287 \\
&192 & 0.250 & 0.311 & 0.269 & 0.328 \\
&336 & 0.315 & 0.356 & 0.325 & 0.366 \\
&720 & 0.415 & 0.405 & 0.421 & 0.415 \\
& Avg. & 0.292 & 0.334 & 0.304 & 0.349 \\
\bottomrule
\end{tabular}
\end{adjustbox}
\vspace{-0.2cm}
\label{tab:fedformer_ltsf}
\end{table}

\subsection{Predictor diversity}\label{adx:pred_diversity}

To bolster our claim that the proposed Predictor-Corrector framework is agnostic to the underlying Predictor, we evaluate our Corrector on two additional Predictors, i.e., RNN and TCN. The datasets are \textsc{Lorenz} and \textsc{FHN}. The extrapolation columns list the horizon up to which the Corrector brings at least 3\% reduction in MSE of Predictor. The results are shown in the Table. \ref{tab:rnn_tcn_table}.

\begin{table}[H]
\centering
\small 
\caption{Test MSE of RNN \& TCN as Predictors on Lorenz \& FHN dataset with (w/) and without (w/o) Corrector. The $\% \downarrow$ shows the percentage reduction in MSE of RNN \& TCN with our proposed Corrector. For both interpolation and extrapolation, reported MSE values are computed from timestep 0 up to the specified timestep ($t$) for each setting (0-$t$).}
\vspace{0.05cm}
\renewcommand{\arraystretch}{1.2}
\setlength{\tabcolsep}{4pt} 
\begin{adjustbox}{max width=\textwidth} 
\begin{tabular}{c|c|cc|cc}
\toprule
\multicolumn{1}{c|}{\multirow{2}{*}{\parbox[c]{2cm}{\centering \textbf{Dynamical}\\ \textbf{System}}}} & 
\multicolumn{1}{c|}{\multirow{2}{*}{\textbf{Model}}} &
\multicolumn{2}{c|}{\multirow{1}{*}{\textbf{Interpolation}}} &
\multicolumn{2}{c}{\multirow{1}{*}{\textbf{Extrapolation}}} \\
\cline{3-6}
& & RNN & TCN & RNN & TCN \\  
\midrule
\multicolumn{1}{c}{\multirow{2}{*}{\parbox[c]{2cm}{\centering Lorenz}}}
& \multicolumn{1}{|c|}{w/o}
& $1.102$ & $0.725$ & $5.705$ & $3.908$ \\
\cline{2-6}
& \multicolumn{1}{|c|}{w/}
& $0.901$ & $0.635$ & $5.504$ & $3.709$ \\
\cline{2-6}
& \multicolumn{1}{|c|}{$0$ -- $t$ $\boldsymbol{|}$ \%$\downarrow$} 
& $0$--$50$ $\boldsymbol{|}$ $18\%$ & $0$--$50$ $\boldsymbol{|}$ $13\%$ & $0$--$80$ $\boldsymbol{|}$ $3\%$ & $0$--$100$ $\boldsymbol{|}$ $5\%$  \\
\midrule 
\multicolumn{1}{c}{\multirow{2}{*}{\parbox[c]{2cm}{\centering Glycolytic \\ Oscillator}}}
& \multicolumn{1}{|c|}{w/o}
& $0.190$ & $0.140$ & $0.187$ & $0.160$ \\
\cline{2-6}
& \multicolumn{1}{|c|}{w/}
& $0.140$ & $0.110$ & $0.180$ & $0.150$ \\
\cline{2-6}
& \multicolumn{1}{|c|}{$0$ -- $t$ $\boldsymbol{|}$ \%$\downarrow$} 
& $0$--$50$ $\boldsymbol{|}$ $26\%$ & $0$--$50$ $\boldsymbol{|}$ $21\%$ & $0$--$90$ $\boldsymbol{|}$ $4\%$ & $0$--$110$ $\boldsymbol{|}$ $6\%$ \\
\bottomrule
\end{tabular}
\end{adjustbox}
\label{tab:rnn_tcn_table}
\end{table}

\newpage
\section{Performance of Predictor-Corrector on Exchange (8D)}
\begin{figure}[H]
    \centering
    \includegraphics[width=0.60\textwidth]{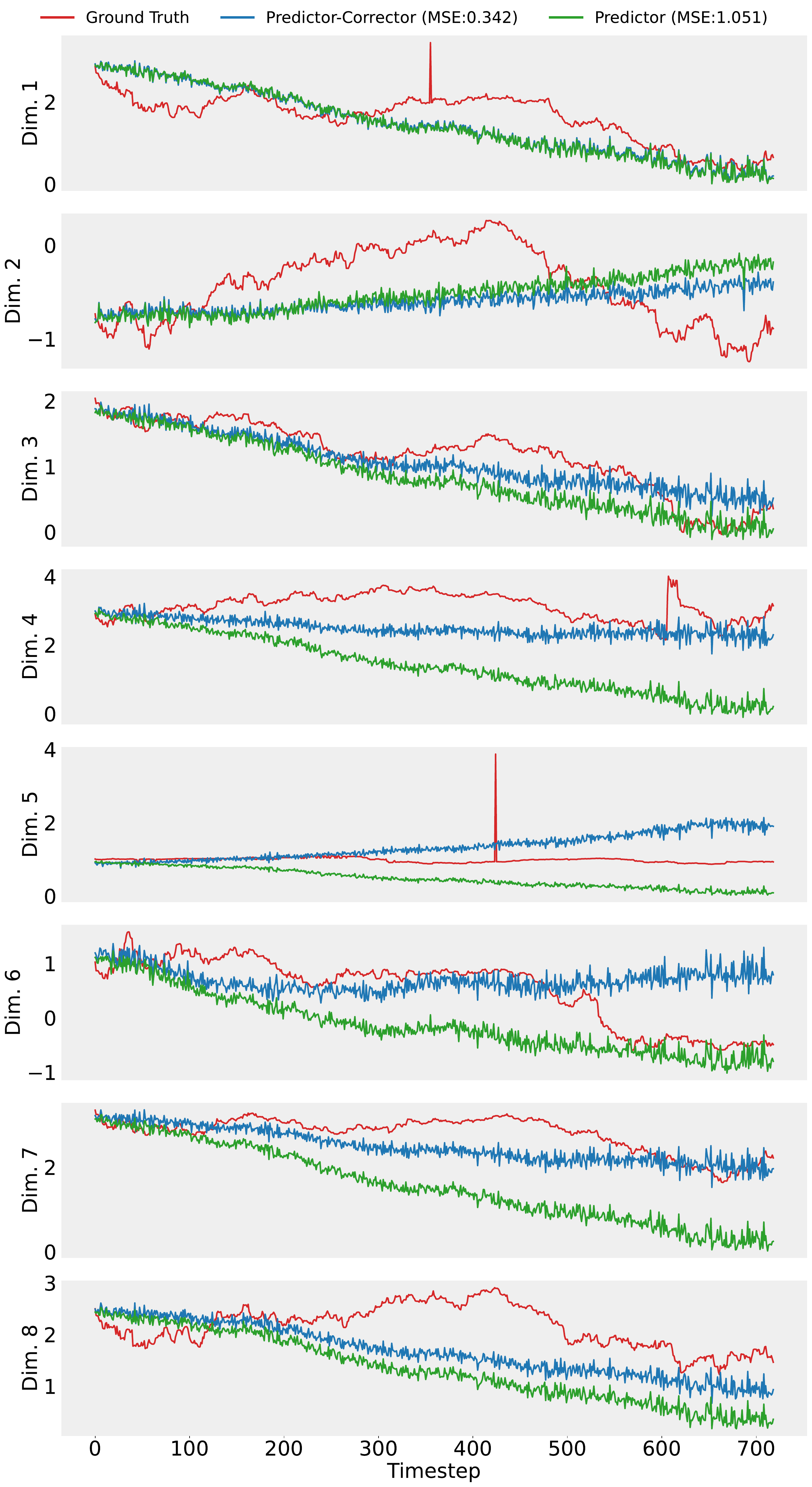}
    \caption{The performance of Corrector on one of the trajectories of the Exchange test dataset. Dim. is used as a shorthand for dimension in the plots.}
    \label{fig:exchange_data}
\end{figure}

Table \ref{tab:forecast_results} shows the performance of DLinear without Corrector (w/o) and with Corrector (w/). Fig. \ref{fig:exchange_data} shows the performance of Predictor-Corrector compared to Predictor alone on one of the trajectories from the Exchange test dataset. The trajectory shows a drop in MSE with Corrector. 

\section{LTSF dataset statistics}\label{adx:forecast_dataset}
Table~\ref{tab:forecast_dataset} lists the statistics of the five LTSF datasets used in Section~\ref{sec:fore_data}, including the number of variates, total timesteps, and sampling granularity.

\begin{table}[t]
    \centering
    \caption{The statistics of five datasets from Table \ref{tab:forecast_results}} 
    \vspace{0.15cm}
    \scalebox{0.70}{%
    \begin{tabular}{c|c|c|c|c|c}
    \toprule
         Datasets & Exchange & ETTm2 & ETTh2 & ILI & Weather \\ 
        \cline{1-6}
        Variates & 8 & 7 & 7 & 7 & 21 \\
        Timesteps & 7,588 & 69,680 & 17,420 & 966 & 52,696 \\
        Granularity & 1 day & 15 min & 1 hour & 1 week & 10 min \\
        \bottomrule
    \end{tabular}%
    }
    \vspace{-10pt}
    \label{tab:forecast_dataset}
\end{table}

\section{Performance of Predictor-Corrector on \textsc{Pen} (45D)}\label{adx:pen_viz}
\begin{figure}[H]
    \centering
    \includegraphics[width=0.65\textwidth]{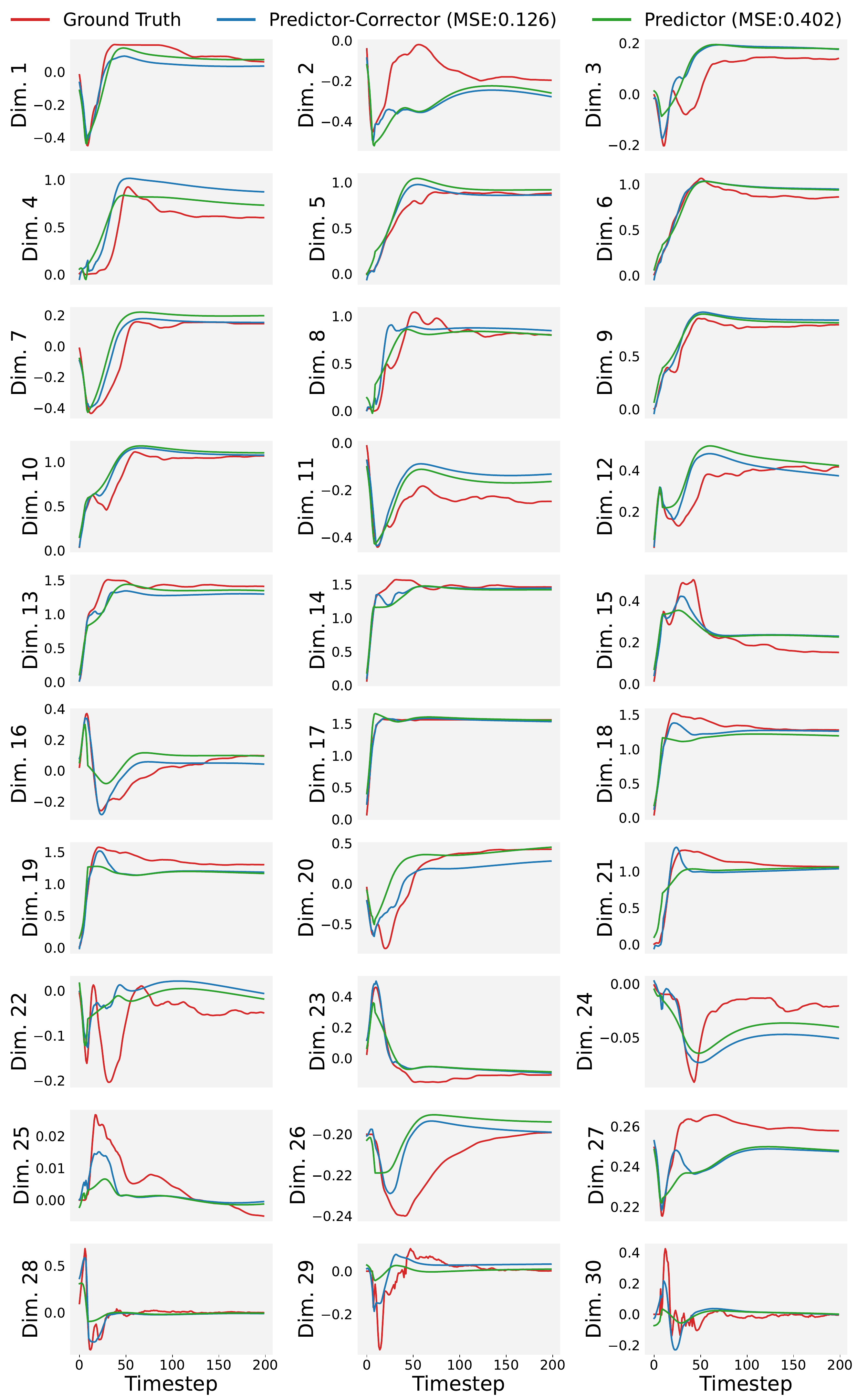}
    \caption{The performance of Corrector on one of the trajectories of \textsc{Pen} for a 20\% observed points setting. The first 30 dimensions of the \textsc{Pen} trajectory are shown here. Dim. stands for dimension.}
    \label{fig:pen_first30}
\end{figure}

\begin{figure}[H]
    \centering
    \includegraphics[width=0.75\textwidth]{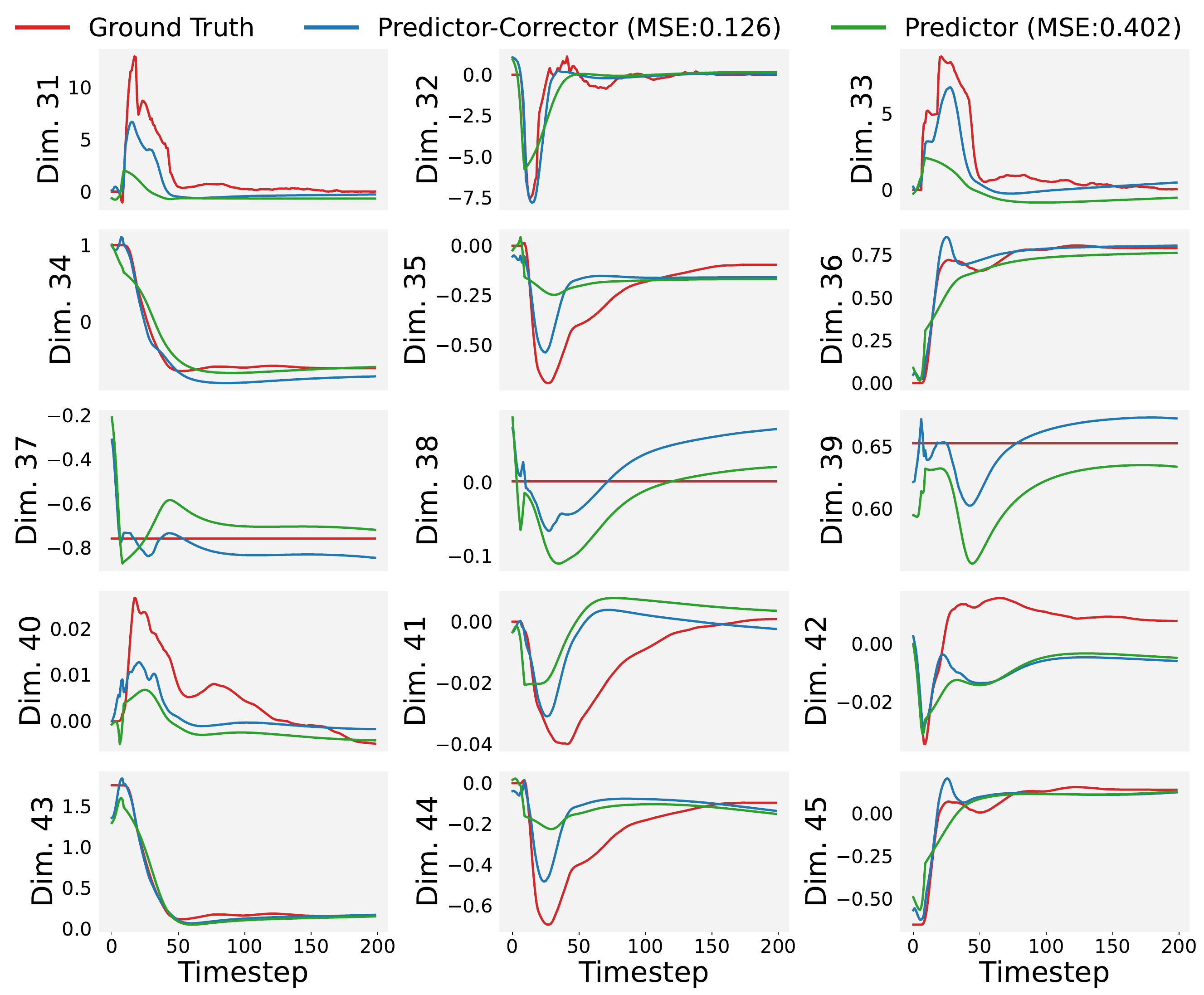}
    \caption{The performance of Corrector on one of the trajectories of \textsc{Pen} for a 20\% observed points setting. The last 15 dimensions of the \textsc{Pen} trajectory are shown here. Dim. stands for dimension.}
    \label{fig:pen_last15}
\end{figure}

The performance of the ContiFormer without Corrector (w/o) and with Corrector (w/) on \textsc{Pen} is shown in Table \ref{tab:walker2d_pen}. Here, we show the performance of Corrector on one of the trajectories from the test dataset for a 20\% observed points setting. The first 24 dimensions are shown in Fig. \ref{fig:pen_first30} and the rest of the 21 dimensions in Fig. \ref{fig:pen_last15}. The visualizations demonstrate that the Corrector (trained on the first 50 timesteps) can correct the Predictor up to 200 timesteps, well beyond the training horizon, for such a high-dimensional dynamical system.

\end{document}